\pdfoutput=1
\documentclass[11pt]{article}

\usepackage[]{EMNLP2023}

\usepackage{times}
\usepackage{latexsym}
\usepackage{stmaryrd}
\usepackage{amsmath}
\usepackage{multirow}
\usepackage{bbm}
\usepackage{graphicx}
\usepackage{amssymb}
\usepackage{booktabs}
\usepackage{adjustbox}
\usepackage{xspace}
\usepackage{algpseudocode}
\usepackage{algorithm}
\usepackage{graphicx}
\usepackage{caption}
\usepackage{subcaption}
\usepackage{color, colortbl}
\usepackage{xcolor}
\usepackage{tablefootnote}
\usepackage{enumitem}  % More control
\usepackage{lipsum}
\usepackage{xspace}
\usepackage{cleveref}

\usepackage{hyperref}
\def\xHyphenate#1#2\wholeString {\if#1$%
    \else\transform{#1}%
    \takeTheRest#2\ofTheString\fi}
\def\takeTheRest#1\ofTheString\fi
{\fi \xHyphenate#1\wholeString}
\def\transform#1{\url{#1}\hskip 0pt plus 1pt}
\usepackage[T1]{fontenc}
\usepackage[utf8]{inputenc}

\usepackage{microtype}

\usepackage{listings,lstautogobble}
\usepackage{caption}
\usepackage{fancyvrb}
\usepackage{fvextra}
\usepackage{relsize}
\usepackage{tikz}
\usepackage{pgfplots}
\usepackage{tcolorbox}
\usepackage{svg}

\DefineVerbatimEnvironment{MetaVerbatim}
  {Verbatim}
  {formatcom=\color{mvcolor}}% this argument contains options   
\colorlet{mvcolor}{green!60!red}
\newcommand{\tabref}[1]{Table~\ref{#1}\xspace}
\newcommand{\figref}[1]{Figure~\ref{#1}\xspace}

\newcommand{\secref}[1]{Section\xspace\ref{#1}\xspace}
\newcommand{\appref}[1]{Appendix\xspace\ref{#1}\xspace}

\definecolor{color1}{RGB}{141, 211, 199}
\definecolor{color2}{RGB}{255, 255, 179}
\definecolor{color3}{RGB}{190, 186, 218}
\definecolor{color4}{RGB}{251, 128, 114}

\newcommand{\bxbloom}{BX$_\text{BLOOM}$\xspace}
\newcommand{\bxllama}{BX$_\text{LLaMA}$\xspace}

\newcommand{\datasetname}{Bactrian-X\xspace}

\newcommand{\equalsign}{\footnotemark[1]\hspace{0.1cm}}

\newcommand{\sd}[1]{\smaller[2]\ensuremath{\pm#1}}

\usepackage{inconsolata}

\setlist{topsep=1pt,itemsep=1pt,partopsep=1pt, parsep=1pt}

\title{\datasetname: Multilingual Replicable Instruction-Following Models with Low-Rank Adaptation}

\author{
  Haonan Li\textsuperscript{1}\thanks{\hspace{0.2cm}These authors contributed equally.} \quad Fajri Koto\textsuperscript{1}\equalsign \quad Minghao Wu\textsuperscript{1,2} \quad  Alham Fikri Aji\textsuperscript{1} \quad Timothy Baldwin\textsuperscript{1,3}\\
  \textsuperscript{1}Natural Language Processing Department, MBZUAI \\
\textsuperscript{2}Monash University  \textsuperscript{3}The University of Melbourne \\
  \texttt{\{haonan.li,fajri.koto,minghao.wu,alham.fikri,timothy.baldwin\}@mbzuai.ac.ae}
 }

\begin{document}
\maketitle
\begin{abstract}
Instruction tuning has shown great promise in improving the performance of large language models. However, research on multilingual instruction tuning has been limited due to the scarcity of high-quality instruction--response datasets across different languages. To bridge this gap, we present \datasetname, a comprehensive multilingual parallel dataset of 3.4 million instruction--response pairs across 52 languages. Leveraging this dataset, we train a set of adapters using low-rank adaptation (LoRA), which are lightweight components that seamlessly integrate with large language models. These adapters have a substantially lower parameter count than the base model, making them easily replaceable and usable as plug-ins for different languages or language groups. Extensive experiments in various multilingual evaluation settings demonstrate that models derived from LoRA-based training over \datasetname  outperform both the vanilla models and existing instruction-tuned models. The code and models are publicly available at 
%\url{redacted.for.anonymity}.
\url{https://github.com/mbzuai-nlp/bactrian-x}. 

\end{abstract}

\section{Introduction}
\label{sec:intro}

Fine-tuning large language models (LLMs) with instruction--response pair datasets has demonstrated remarkable zero-shot generalization capabilities for open-source and closed-source models~\citep{DBLP:conf/iclr/SanhWRBSACSRDBX22, wei2022finetuned, ouyang2022training, gpt4}. Although the LLMs are often pre-trained using multilingual texts, the instruction-tuning for open-source models is restricted to English \citep{alpaca, vicuna2023, lamini-lm}, bringing into question its multilingual generalizability.  Closed-resource models such as OpenAI GPT-4~\citep{gpt4} and Google BARD,\footnote{\url{https://bard.google.com/}} despite performing impressively over high-resource languages, are still lacking in terms of multilingual generalizability under monolingual instruction tuning.

%Instruction-tuned large language models (LLMs) have demonstrated remarkable zero-shot generalization capabilities~\citep{DBLP:conf/iclr/SanhWRBSACSRDBX22, wei2022finetuned, ouyang2022training, gpt4}. Recent advancements have made instruction-following language models more accessible by fine-tuning smaller models using general instructions from larger language models \citep{alpaca, vicuna2023, lamini-lm}. However, existing work has primarily focused on English.

The scarcity of instruction--response pair datasets in languages beyond English is hinders multilingual instruction tuning. The existing \texttt{xP3} dataset ~\cite{bloomz}, which was used to fine-tune BLOOM and mT5, employs English instructions. Although  \citet{bloomz} also experiments with \texttt{xP3mt} --- machine-translated instructions --- it focuses on classic NLP tasks such as summarization and question answering, rather than general instructions. Additionally, both \texttt{xP3} and \texttt{xP3mt} use template-based prompts, and hence lack variation.

%Research on multilingual instruction-following LLMs is limited, mainly because of the availability of open-source multilingual large foundation models and the scarcity of multilingual instruction datasets. \citet{bloomz}~introduced the xP3 dataset and fine-tuned BLOOM~\citep{bloom} and mT5~\citep{xue-etal-2021-mt5} models on it. However, xP3 is not truly multilingual as the prompts are still in English.  In addition, the dataset primarily focuses on downstream NLP tasks, lacking variation in human-written prompts~\citep{self-instruct}. This results in the limitations of their models in handling general instructions. 

%On the other hand, newer product-oriented LLMs, like OpenAI GPT-4~\citep{gpt4} and Google BARD,\footnote{\url{https://bard.google.com/}} showcase impressive multilingual capabilities in instruction following. However, the closed nature of these models hinders benchmarking and reproducibility, posing challenges to scientific progress in the field.

\begin{figure}[t]
    \centering
    \includegraphics[width=0.9\linewidth]{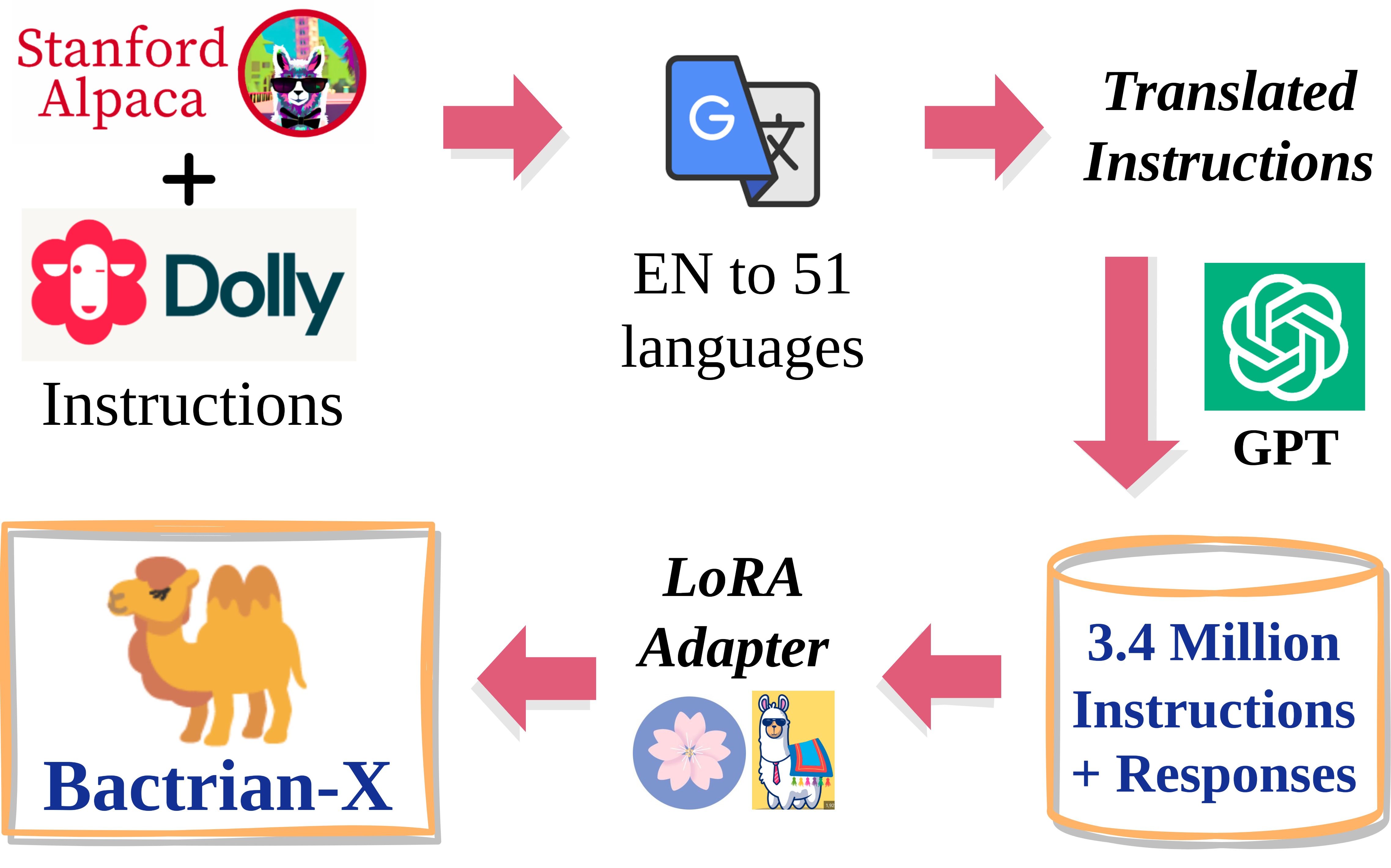}
    \caption{Overview of the \datasetname dataset and process for model creation.}
    \label{fig:bactrian-x}
    \vspace{-0.3cm}
\end{figure}

%To address these gaps, we introduce Bactrian-X, the first-ever publicly available multilingual instruction-following dataset and language model covering 52 languages (see \figref{fig:bactrian-x}).
%Bactrian-X instruction training data is constructed by translating the Alpaca~\citep{alpaca} and Dolly~\citep{dolly} general instruction datasets into 51 languages, followed by distilling ChatGPT outputs from the translated instructions.
%With 67K instruction-response pairs (52K from Alpaca and 15K from Dolly) for each language, the total number of instances reaches 3.4M.
%With 67K instruction-response pairs for each language, the total number of instances reaches 3.4M.
%We employ automatic quality estimation to verify the translation and response quality, and conduct additional human assessment in the responses quality.

To investigate general instruction tuning in a multilingual setting, we introduce \datasetname, containing parallel instruction--response pairs across 52 languages that were automatically constructed by translating instructions from Alpaca~\citep{alpaca} and Dolly~\citep{dolly} via the Google Translate API.\footnote{\url{https://translate.google.com/}} As we detail in \Cref{sec:dataset}, we use the output distillation trick to obtain corresponding responses by leveraging ChatGPT outputs, conditioned on the translated instructions.
%We prefer this way to translation as it allows us in this paper to also assess the quality of ChatGPT response in different languages. Moreover, 
%Translating original responses via the Google Translate API is costly, and using open-source translation such as NLBB \cite{costa2022no} could not warrant a good translation quality in under-represented languages such as Burmese and Tagalog.
% TJB: what is the point of the preceding sentence? We don't use NLBB, right, and Burmese and Tagalog are covered by Google Translate. I have removed it for now
With 67K instruction--response pairs for each language, the total number of instances in Bactrian-X reaches 3.4M.

In contrast to previous multilingual instruction models such as BLOOMZ~\citep{bloomz} which are subject to full fine-tuning via parameter updates across all layers, this study highlights the potential of parameter-efficient fine-tuning techniques, specifically LoRA \citep{hu2022lora}. LoRA uses adapters with substantially fewer parameters than base LLMs, making them more practical and adaptable for real-world applications. Specifically, in this work, we introduce \bxbloom and \bxllama models, which build upon the BLOOM \citep{bloom} and LLaMA \citep{llama} models, and find them to be better than the associated instruction-tuned models: BLOOMZ~\cite{bloomz} and Alpaca~\cite{alpaca}. 
% Through our experiments, we present promising results demonstrating the competitiveness of \bxbloom in comparison to BLOOMZ \cite{bloomz}, which represents BLOOM fully pre-trained on multilingual instruction.

% We conducted thorough evaluations to compare our models with vanilla LLMs and existing instruction-tuned models. The consistent results across various zero-shot multilingual NLP tasks, including \texttt{XCOPA}~\citep{ponti2020xcopa}, \texttt{XStoryCloze}~\citep{lin-etal-2022-shot}, \texttt{XWinograd}~\citep{bloomz}, our own multilingual sentiment analysis dataset \texttt{SentimentX}, and \texttt{EXAMS}~\citep{exams} demonstrate the improved performance of our models compared to all baseline models. Furthermore, we assessed the model's ability to comprehend general instructions using GPT-4 as an evaluator, following the methodology proposed by \citet{vicuna2023}, and further aligned by human evaluation. The results revealed that our models were preferred by both GPT-4 and human annotators over all baseline models.

We conduct a comprehensive series of experiments covering a range of zero-shot multilingual NLP tasks, including \texttt{XCOPA}~\citep{ponti2020xcopa}, \texttt{XStoryCloze}~\citep{lin-etal-2022-shot}, \texttt{XWinograd}~\citep{bloomz}, our own multilingual sentiment analysis dataset \texttt{SentimentX}, and \texttt{EXAMS}~\citep{exams}. The consistently high results across these tasks highlight the effectiveness of our multilingual instruction dataset and adapter technique for instruction tuning in languages beyond English. To further validate our findings, we use GPT-4 as an evaluator based on the methodology proposed by \citet{vicuna2023}, and additionally conduct human evaluation with native speakers. All results confirm that our proposed models outperform the vanilla foundation models and existing instruction-tuned models.

%%% Local Variables:
%%% mode: latex
%%% TeX-master: "emnlp2023"
%%% End:

\section{Related Work}

\paragraph{Multilingual Instruction Tuning}
LLMs such as GPT-3 \cite{NEURIPS2020_1457c0d6}, PaLM \cite{DBLP:journals/corr/abs-2204-02311} and LLaMA \cite{llama} \cite{DBLP:journals/corr/abs-2203-15556, bloom, zeng2023glmb} have revolutionized NLP.
Research has demonstrated that fine-tuning LLMs with instruction prompts can improve their capacity to perform unseen/novel tasks \cite{wei2022finetuned,DBLP:conf/iclr/SanhWRBSACSRDBX22,ouyang2022training,DBLP:journals/corr/abs-2210-11416,bloomz}.
Recently, \citet{self-instruct,alpaca} showed that machine-generated instructions can be used for instruction tuning. 
%\citet{} generate a set of 52K instructions to fine-tune LLaMA and demonstrate performance gain.
\citet{lamini-lm} created a large-scale dataset with 2.6M instructions, and demonstrated that relatively small language models also benefit from the instructions.
%Furthermore, a large number of prior works have demonstrated that unified multilingual models are more resource-efficient and promote effective knowledge transfer \cite{pires-etal-2019-multilingual, aharoni-etal-2019-massively, conneau-etal-2020-unsupervised, aji-etal-2020-neural, xue-etal-2021-mt5, wu-etal-2021-uncertainty, li-etal-2022-universal, aji-etal-2022-one}.
Prior work has predominantly been on English, and instruction-tuning in languages beyond English remains limited.
%However, there are few multilingual instruction-tuned LLMs due to the unavailability of the pretrained LLMs and multilingual instruction datasets.
The closest work to ours is BLOOMZ~\cite{bloomz}, which finetunes BLOOM~\cite{bloom} and mT5~\cite{xue-etal-2021-mt5} on the \texttt{xP3} and \texttt{xP3mt} multilingual instruction datasets. However, both \texttt{xP3} and \texttt{xP3mt} are based on human-written templates, and lack the variability of an organic multilingual dataset. %
Our work, instead, constructs a parallel general instruction dataset by translating English instructions into 51 languages and generating responses via ChatGPT \cite{ouyang2022training}. To the best of our knowledge, our Bactrian-X instruction dataset is the largest general-purpose multilingual instruction dataset to date.

\begin{table*}[t]
\centering
\small
\begin{tabular}{lrlr@{}lr@{}lr@{}lr@{}l}
\toprule
Tokenizer &Vocab size & Lang & \multicolumn{2}{c}{Instruction tokens} & \multicolumn{2}{c}{Input tokens} & \multicolumn{2}{c}{Response tokens} & \multicolumn{2}{c}{Total tokens} \\
\midrule
mBART-50 & 250,054 & all & 17.11 & \sd{1.96} & 27.54 & \sd{2.84} & 133.65 & \sd{17.4} & 178.30 & \sd{22.2} \\[0.7ex]

\multirow{2}{*}{BLOOM} &\multirow{2}{*}{251,680} & seen & 16.14 &\sd{2.87} & 25.98 &\sd{3.99} & 128.88 &\sd{25.5} & 171.00 & \sd{31.3} \\
  && unseen & 34.21 &\sd{22.0} & 51.41 &\sd{31.7} & 275.97 &\sd{179} & 361.60 & \sd{231} \\[0.7ex]

\multirow{2}{*}{LLaMA} & \multirow{2}{*}{32,000} & seen & 23.13 & \sd{2.78} & 36.69 & \sd{3.85} & 185.18 & \sd{18.2} & 244.96 & \sd{24.3} \\
  && unseen & 57.22 & \sd{35.6} & 86.93 & \sd{52.6} & 448.61 & \sd{293} & 592.77 & \sd{376} \\
\bottomrule
\end{tabular}
\caption{
Average \# tokens in each Instruction, Input, and Response across all
languages. Note that the token counts for mBART-50, LLaMA, and BLOOM are
based on the respective tokenizers and are not directly comparable.
mBART-50 covers all 52 languages, while LLaMA and BLOOM cover only a
subset of the languages in Bactrian-X, and separate results are thus presented
for seen and unseen languages.
}
\label{tab:data_stat}
\end{table*}

%%% Local Variables:
%%% mode: latex
%%% TeX-master: "../emnlp2023"
%%% End:

\paragraph{Parameter Efficient Fine-Tuning (PEFT)}
%LLMs commonly require massive computational resources for both training and inference.
%Fine-tuning all parameters, as seen in Alpaca \cite{alpaca}, Vicuna \cite{vicuna2023} and LaMini-LM \cite{lamini-lm}, can be computationally expensive.
%To mitigate this, researchers proposed Parameter Efficient Fine-Tuning (PEFT). 
Fine-tuning all parameters of an LLM (e.g.~Alpaca \cite{alpaca}, Vicuna \cite{vicuna2023} and LaMini-LM \cite{lamini-lm}) is computationally expensive, and adapters \cite{pmlr-v97-houlsby19a} offer a more cost-effective alternative.
PEFT updates a small number of parameters during fine-tuning, and achieves comparable performance to fully fine-tuned counterparts \cite{pmlr-v97-houlsby19a, guo-etal-2021-parameter, lester-etal-2021-power, ben-zaken-etal-2022-bitfit}. \citet{hu2022lora} introduced Low-Rank Adaptation (LoRA), which incorporates trainable rank decomposition matrices into transformer layers \cite{vaswani2017attention} during fine-tuning without introducing additional latency during inference. They demonstrate that by fine-tuning with less than 1\% of the model parameters, LoRA outperforms several fully fine-tuned LLMs, including GPT-3 \cite{NEURIPS2020_1457c0d6}, on various tasks.

In recent work, \citet{alpaca} use the LoRA trick to fine-tune LLaMA~\cite{llama}, resulting in the Alpaca model, but did not carry out comprehensive evaluation. In this work, we also leverage the LoRA technique to develop a range of monolingual and multilingual adapters, with a much larger instruction--response dataset, across 52 languages. We provide empirical analysis based on automatic and human evaluation to demonstrate the effectiveness of our method.

%serving diverse use cases while simultaneously reducing computational costs and minimizing the carbon footprint during training.

%%% Local Variables:
%%% mode: latex
%%% TeX-master: "emnlp2023"
%%% End:

\section{\datasetname Dataset}
\label{sec:dataset}

In this section, we detail the dataset creation process and provide an overview of the resulting data, focusing on the quality of translated instructions and generated responses.

\subsection{Dataset Creation}
We construct the \datasetname dataset in two steps: instruction translation, and response generation (see \figref{fig:bactrian-x}). 

\paragraph{Instruction Translation}
We use English instructions developed for  Alpaca (52K) and Dolly (15K), and use the Google Translate API to translate them into 51 different languages, based on the languages used for mBART-50 \cite{tang2020multilingual}. The Alpaca instructions were automatically generated by GPT-3.5~\cite{ouyang2022training}
via the self-instruct technique \cite{self-instruct}, while the Dolly dataset was manually curated by thousands of Databricks company employees. Prior to the translation process, we identify instructions containing programming-related content based on a keyword-matching method and exclude them from the translation process. The total cost for translating the instructions was approximately USD\$10,000.

\paragraph{Response Generation}
For each translated instruction, we use ChatGPT (\texttt{gpt-3.5-turbo}) to obtain a response.\footnote{The response generation was conducted during April 16--21, 2023.} For English, we pair the instruction with the original response. Translating responses into the 51 languages is costly. Moreover, potential issues such as ``translationese'' and non-native answer styles may arise from relying solely on translated responses. The total cost for generating responses amounts to around \$3,000 USD. We leave the comparison between the translated responses and the ChatGPT-generated responses to future work. 

%Upon translating the English instructions into multiple languages, we proceed to generate corresponding responses for each instruction using the ChatGPT (\texttt{gpt-3.5-turbo}). 
%While the original English instructions are paired with responses, which means the same translation can be conducted on these responses, we opt to generate responses directly from ChatGPT.\footnote{The generation was conducted during April 16-21, 2023.} 
%This decision is motivated by the desire to avoid potential issues such as ``translationese'' and non-native answer styles that may arise from relying solely on translated responses.
%The total cost for generating responses amounts to around \$3,000 USD. We leave the comparison between the translated responses and the generated responses (ours) to future work. 

\begin{table}[t]
\small
\centering
\begin{tabular}{@{}lccc@{}}
\toprule
        & BLEU & chrF++ & COMET \\ \midrule
min     & 28.0 & 52.5   & 82.3  \\
25\% Q. & 42.9 & 64.7   & 88.7  \\
mean    & 48.1 & 68.1   & 90.2  \\
% median  & 47.2 & 67.8   & 90.5  \\
75\% Q. & 52.7 & 72.2   & 92.0  \\
max     & 69.0 & 82.7   & 95.3  \\ \bottomrule
\end{tabular}
\caption{
    BLEU, chrF++ and COMET scores for the language pairs from the 51 languages to English.
    COMET scores are up-scaled by $\times 100$.
}
\label{tab:mt_quality}
\end{table}

\subsection{Exploratory Data Analysis}
\label{sec:data}

\paragraph{Dataset Statistics}

%We calculate the overall statistics of tokenized texts in the 52 languages based on mBART-50, LLaMA, and BLOOM tokenizers, and present the results in \tabref{tab:data_stat}. The 52 languages have been seen by mBART-50, thus the average number of tokens is relatively smaller. For languages unseen by BLOOM and LLaMA, the tokenized texts are 2--3 times longer than mBART-50.
%This implies that for the unseen languages, both BLOOM and LLaMA require a larger sequence length, making it more challenging to effectively adapt them with LoRA adapter.

We analyzed the tokenized texts in the 52 languages using the mBART-50, LLaMA, and BLOOM tokenizers, and present the statistics in \tabref{tab:data_stat}. Since mBART-50 is trained on all 52 languages, the tokenizer is trained on all the languages, and the average number of tokens is thus relatively smaller than LLaMA and BLOOM. However, for languages unseen by BLOOM and LLaMA, the tokenized texts are 2 to 3 times longer compared to mBART-50. This suggests that for these unseen languages, both BLOOM and LLaMA models require a larger sequence length for semantically similar input texts, posing a challenge for effective adaptation with the LoRA adapter.

%We conducted text tokenization on instructions, inputs, and responses across all 52 languages using different tokenizers including mBART-50, LLaMA, and BLOOM, and present the overall statistics in \tabref{tab:data_stat}.
%For the mBART-50 tokenizer, we provide the average number of tokens across all 52 languages, as these languages are fully covered by the tokenizer.
%Regarding the LLaMA and BLOOM tokenizers, their supported languages partially overlap with our dataset. Therefore, we present statistics for both the languages seen and unseen by the tokenizers.
%From the table, we observe that for languages seen by the tokenizers, the BLOOM tokenizer yields a similar number of tokens as the mBART-50 tokenizer. However, the LLaMA tokenizer produces more tokens, indicating that the text is divided into more sub-word units, resulting in longer sequences.
%For unseen languages, the LLaMA tokenizer results in an average of 593 total tokens, which is three times the number of tokens produced by the mBART-50 tokenizer.  This implies that not only does the model require a larger sequence length to accommodate these languages, but it also faces greater challenges in effectively adapting to them.

\begin{figure}[t]
    \centering
    \begin{subfigure}[b]{0.9\linewidth}
        \centering
        \begin{tikzpicture}
            \begin{axis}[
                ybar stacked,
                height=4.5cm,
                width=1\linewidth,
                bar width=15pt,
                point meta=explicit symbolic,
                nodes near coords,
                every node near coord/.append style={font=\footnotesize},
                tick label style={font=\footnotesize},
                enlargelimits=0.15,
                legend style={
                    at={(0.5,-0.3)},
                    anchor=north,
                    legend columns=-1
                },
                ylabel={Proportion (\%)},
                ylabel style={yshift=-0.2cm},
                ymin=0, ymax=1,
                symbolic x coords={0, 1, 2, 3, 4, 5},
                xticklabels={ar,id,zh,my,ta,tl},
                xticklabel style={text height=2ex,
                    yshift=+0.1cm,},
                xtick=data,
                ]
            \addplot+[ybar, fill=color1, draw=color1] plot coordinates {(0,0.5595854922279793) (1,0.9396984924623115) (2,0.965) (3,0.04040404040404041) (4,0.1407035175879397) (5,0.25263157894736843)};
            \addplot+[ybar, fill=color2, draw=color2] plot coordinates {(0,0.37305699481865284) (1,0.04522613065326633) (2,0.005) (3,0.13131313131313133) (4,0.2814070351758794) (5,0.37894736842105264)};
            \addplot+[ybar, fill=color3, draw=color3] plot coordinates {(0,0.06217616580310881) (1,0.010050251256281407) (2,0.0) (3,0.5555555555555556) (4,0.41708542713567837) (5,0.2)};
            \addplot+[ybar, fill=color4, draw=color4] plot coordinates {(0,0.0051813471502590676) (1,0.005025125628140704) (2,0.03) (3,0.2727272727272727) (4,0.16080402010050251) (5,0.16842105263157894)};
            \legend{\texttt{Rate-A}, \texttt{Rate-B}, \texttt{Rate-C}, \texttt{Rate-D}}
            \end{axis}
        \end{tikzpicture}
        \caption{
            Human evaluation of response fluency. 
        }
        \label{fig:fluency}
    \end{subfigure}
    \\[.4cm]
    \begin{subfigure}[b]{0.9\linewidth}
        \centering
        \begin{tikzpicture}
            \begin{axis}[
                ybar stacked,
                height=4.5cm,
                width=1\linewidth,
                bar width=15pt,
                point meta=explicit symbolic,
                nodes near coords,
                every node near coord/.append style={font=\footnotesize},
                tick label style={font=\footnotesize},
                enlargelimits=0.15,
                legend style={
                    at={(0.5,-0.3)},
                    anchor=north,
                    legend columns=-1
                },
                ylabel={Proportion (\%)},
                ylabel style={yshift=-0.2cm},
                ymin=0, ymax=1,
                symbolic x coords={0, 1, 2, 3, 4, 5},
                xticklabels={ar,id,zh,my,ta,tl},
                xtick=data,
                xticklabel style={text height=2ex,
                    yshift=+0.1cm,},
                ]
            \addplot+[ybar, fill=color1, draw=color1] plot coordinates {(0,0.5347593582887701)  (1,0.7894736842105263) (2,0.91) (3,0.055865921787709494)  (4,0.06914893617021277) (5,0.5591397849462365) };
            \addplot+[ybar, fill=color2, draw=color2] plot coordinates {(0,0.3315508021390374) (1,0.13157894736842105) (2,0.045) (3,0.16759776536312848) (4,0.09574468085106383) (5,0.2903225806451613)};
            \addplot+[ybar, fill=color3, draw=color3] plot coordinates {(0,0.10695187165775401) (1,0.06842105263157895) (2,0.02) (3,0.3575418994413408) (4,0.30319148936170215) (5,0.0967741935483871)};
            \addplot+[ybar, fill=color4, draw=color4] plot coordinates {(0,0.026737967914438502) (1,0.010526315789473684) (2,0.025) (3,0.41899441340782123) (4,0.5319148936170213) (5,0.053763440860215055)};
            \legend{\texttt{Rate-A}, \texttt{Rate-B}, \texttt{Rate-C}, \texttt{Rate-D}}
            \end{axis}
        \end{tikzpicture}
        \caption{
            Human evaluation of response informativeness. 
        }
        \label{fig:informativeness}
    \end{subfigure}
    \caption{
        Human evaluation of the response quality for Bactrian-X. Rate A is the best and D is the worst.
    }
    \label{fig:response_quality}
\end{figure}
%%% Local Variables:
%%% mode: latex
%%% TeX-master: "../emnlp2023"
%%% End:

\paragraph{Instruction Quality}
%Because all the instructions are translated by Google Translate API, it is of concern if the instructions are correctly translated into other languages.
%We randomly sample 100 sentences for each language, have the sampled sentences back-translated into English using Google Translate API, and use the corresponding English sentences from the original English dataset as the references to measure the translation quality.
%We use the lexical metric BLEU \cite{papineni-etal-2002-bleu, post-2018-call},\footnote{\texttt{nrefs:1|case:mixed|eff:no|tok:13a|smooth:exp|\\version:2.3.1}} chrF++ \cite{popovic-2017-chrf},\footnote{\texttt{nrefs:1|case:mixed|eff:yes|nc:6|nw:2|\\space:no|version:2.3.1}} and the neural metric COMET \cite{rei-etal-2020-comet},\footnote{\texttt{Unbabel/wmt22-comet-da}} and report the results in \tabref{tab:mt_quality}.
%The worst BLEU score of 28 is on Mongolian-English translation and the BLEU scores for most of the language pairs are higher than 40, which indicates that Google Translate API can produce high-quality and reliable translations when we translate the English instructions into all the other foreign languages.

To test the quality of the translated instructions, we verified the quality of 100 randomly-sampled instances for each language by performing back-translation into English using the Google Translate API. We evaluate the quality of the back-translated instructions relative to the originals based on BLEU \cite{papineni-etal-2002-bleu, post-2018-call},\footnote{\texttt{nrefs:1|case:mixed|eff:no|tok:13a|smooth:exp|\\version:2.3.1}} chrF++ \cite{popovic-2017-chrf},\footnote{\texttt{nrefs:1|case:mixed|eff:yes|nc:6|nw:2|\\space:no|version:2.3.1}} and the trained metric COMET \cite{rei-etal-2020-comet}.\footnote{\texttt{Unbabel/wmt22-comet-da}} The worst BLEU score of 28 is for Mongolian--English translation, but as seen in \tabref{tab:mt_quality}, most language pairs achieved BLEU scores above 40, indicating high quality and reliability of the \datasetname instructions.

\paragraph{Response Quality}
To evaluate response quality, we conducted human evaluations in three high-resource languages --- Arabic (\texttt{ar}), Indonesian (\texttt{id}), Chinese (\texttt{zh}) --- and three low-resource languages --- Burmese (\texttt{my}), Tamil (\texttt{ta}), and Tagalog (\texttt{tl}).
For each language, two native-speaker annotators are asked to assess the fluency and informativeness of the responses given the question, except Tagalog, which had only one annotator.
The quality assessment guideline is provided in \appref{app:quality_guideline}, and the results are shown in \figref{fig:response_quality}, with an inter-annotator agreement (IAA) averaged by language of 0.70 and 0.69 for fluency and informativeness, respectively. 
The results showed that high-resource languages consistently achieved over 80\% satisfactory ratings (A and B), while some low-resource languages like Tamil and Burmese had a significant proportion of lower ratings (C and D). 
This suggests that the outputs generated by ChatGPT are lacking for some low-resource languages. We leave the improvement of data quality for low-resource languages to future work.

%%% Local Variables:
%%% mode: latex
%%% TeX-master: "emnlp2023"
%%% End:

\section{Bactrain-X Models}

%We trained several multilingual and monolingual adapters using LLaMA \cite{llama} and BLOOM \cite{bloom} as foundation models. 
%Considering the limitations of computation resources, we selected the models in 7B or 13B parameters.
%For the monolingual models, we trained 15 LLaMA-7B-based models and 18 BLOOM-7B-based models. The languages were selected if it is present during the pretraining of the foundation models.
%In addition, we trained three multilingual models: Bactrian-X$_{llama}$ (7B, 13B), and Bactrian-X$_{bloom}$ (7B). These models utilize the full dataset encompassing 52 languages. All models are publicly available in our model repository.

Given limitations of computation resources, we use base LLMs with 7B and 13B parameters only. First, we trained three multilingual Bactrian models (BX) over the parallel dataset in 52 languages: \bxllama (7B, 13B), and \bxbloom (7B).\footnote{We do not train \bxbloom (13B) because BLOOM (13B) is not available.} While our primary results are based on the BX models, we additionally train some 7B monolingual Bactrian models (BM) for analysis in Section~\ref{sec:nlp_eval}: 14 BM$_\text{LLaMA}$ and 18 BM$_\text{BLOOM}$. All models will be made publicly available in our model repository.

\begin{table*}[t]
\small
\centering
\begin{tabular}{@{}lccccc@{}}
\toprule
Model & XCOPA & XStoryCloze & XWinograd & SentimentX & EXAMS\\
\midrule
LLaMA (7B) & 50.22 & 57.03 & 57.96 & 30.98 & 28.20 \\
Alpaca-LoRA (7B) & 50.25 & 56.75 & 57.70 & 35.03 & 28.82 \\
\bxllama (7B) & \textbf{51.76} & \textbf{58.91} & \textbf{60.16} & \textbf{42.65} & \textbf{29.14} \\
\midrule
LLaMA (13B) & 51.04 & 57.88 & 52.97 & 33.52 & 30.41 \\
Alpaca-LoRA (13B) & \textbf{54.82} & 59.03 & 52.27 & 35.79 & 30.47 \\ 
\bxllama (13B) & 53.27 & \textbf{62.12} & \textbf{63.65} & \textbf{50.27} & \textbf{35.71} \\
\midrule
BLOOM (7B) & 51.95 & 56.53 & 57.97 & 26.88 & 25.06 \\ 
BLOOMZ (7B) & 52.13 & 58.05 & 60.05 & \textbf{37.68} & \textbf{31.23} \\
\bxbloom (7B) & \textbf{54.78} & \textbf{58.56} & \textbf{60.83} & 33.28 & 26.20 \\
\bottomrule
\end{tabular}
\caption{
    Zero-shot experiment results on downstream tasks. We report averaged accuracy for \texttt{XCOPA}, \texttt{XStoryCloze}, \texttt{XWinograd}, and \texttt{EXAMS}, and macro-F1 scores for \texttt{SentimentX}.
}
\label{tab:experiment_downstream}
\end{table*}

%%% Local Variables:
%%% mode: latex
%%% TeX-master: "../emnlp2023"
%%% End:

We train our LoRA adapters \citep{hu2022lora} using PyTorch with the HuggingFace PEFT implementation \citep{peft,wolf-etal-2020-transformers}.
Hyperparameters used for training the different models can be found in \appref{sec:hyper_parameters} (Table~\ref{tab:hyper_parameters}).
In our evaluation, we compare each multilingual BX model with: (1) the corresponding vanilla models, and (2) the instruction-tuned models Alpaca \cite{alpaca} and BLOOMZ \cite{bloomz}. Details of these models are provided in \appref{app:models}.

%In our preliminary evaluation, we focused on evaluating three multilingual adapters

%%% Local Variables:
%%% mode: latex
%%% TeX-master: "emnlp2023"
%%% End:

\section{Evaluation on NLP Benchmarks}\label{sec:nlp_eval}

In order to thoroughly evaluate our Bactrian-X models, we conducted experiments on various multilingual downstream NLP tasks. 
We first introduce the benchmark datasets we used, and then present the evaluation results in two categories: language understanding tasks (\secref{sec:nlu_results}) and knowledge-intensive tasks (\secref{sec:exam_results}).

\subsection{Datasets}\label{sec:nlp_testsets}
To probe the zero-shot language understanding capability of the different models, we evaluate on the following test sets:
\begin{itemize}
\item \texttt{XCOPA} \cite{ponti2020xcopa}: a multilingual resource designed for causal commonsense reasoning, encompassing 11 languages. The task involves predicting the correct next sentence from two options based on cause and effect question types.

\item \texttt{XStoryCloze} \cite{lin-etal-2022-shot}: a translation of the English story cloze dataset \cite{mostafazadeh-etal-2016-corpus} into 10 languages. The objective is to select one sentence as a plausible ending (closure) from two options, given a four-sentence story as the premise.

\item \texttt{XWinoGrad} \cite{tikhonov-ryabinin-2021-heads,bloomz}: a multilingual benchmark for commonsense reasoning, made up of Winograd Schema Challenge problems in six languages.\footnote{\url{https://cs.nyu.edu/~davise/papers/WinogradSchemas/WS.html}} 
The task involves selecting the most plausible sentence from options that differ slightly.

\item \texttt{SentimentX}: a sentiment classification dataset comprising 3-way sentiment labels collected from various sources, in the following languages: Arabic (ar) \cite{alturayeif-etal-2022-mawqif}, Spanish (es),\footnote{\url{http://tass.sepln.org/2020/}} Japanese (jp) \cite{hayashibe-2020-japanese}, Russian (ru),\footnote{\url{https://github.com/antongolubev5/Russian-Sentiment-Analysis-Evaluation-Datasets}} Indonesian (id) \cite{koto-etal-2020-indolem}, Javanese (jav) \cite{winata-etal-2023-nusax}, Sundanese (sun)  \cite{winata-etal-2023-nusax}, and Swahili (sw) \cite{muhammadSemEval2023}.
\end{itemize}
We also measure how much knowledge the model encodes using the \texttt{EXAMS} benchmark:
\begin{itemize}
\item \texttt{EXAMS} \cite{exams}: a multilingual question-answering dataset made up of multiple-choice questions from high school examinations in 16 languages. It covers subjects from natural science (e.g., physics), social science (e.g., history), to humanities (e.g., philosophy). 
Given that all our experiments are zero-shot, we merge the train, validation, and test sets into a single evaluation dataset, and exclude questions without four multiple choice options, resulting in a total of 20,559 questions.
\end{itemize}

\subsection{Language Understanding Tasks}\label{sec:nlu_results}

The average performance across all languages for \texttt{XCOPA}, \texttt{XStoryCloze}, \texttt{XWinograd}, and \texttt{SentimentX} is presented in Table \ref{tab:experiment_downstream}. During inference, we use translated prompts and sentiment labels in the respective languages, obtained from the Google Translate API. We observe that integrating LoRA with the base models of LLaMA and BLOOM, and training over the multilingual instruction datasets, consistently improves performance over the base models. 
Improvements can also be observed over existing instruction-tuned models such as Alpaca-LoRA, on most tasks. 
For the larger models, we observe further enhancements again, as seen for \bxllama (13B) over LLaMA (13B). 

From the third block, we observe that \bxbloom performs better than the full fine-tuned BLOOMZ model on three out of five tasks. 
Although the performance difference is relatively small, it is worth noting that \bxbloom is fine-tuned only using the LoRA adapter on a smaller multilingual dataset (2.5M samples), whereas BLOOMZ is fully fine-tuned using a larger dataset of 78M samples. Additionally, BLOOMZ is fine-tuned on xP3, which is designed to handle NLP downstream tasks, while \datasetname is more general purpose.

\begin{figure}[t]
	\centering
	\includegraphics[width=1\linewidth]{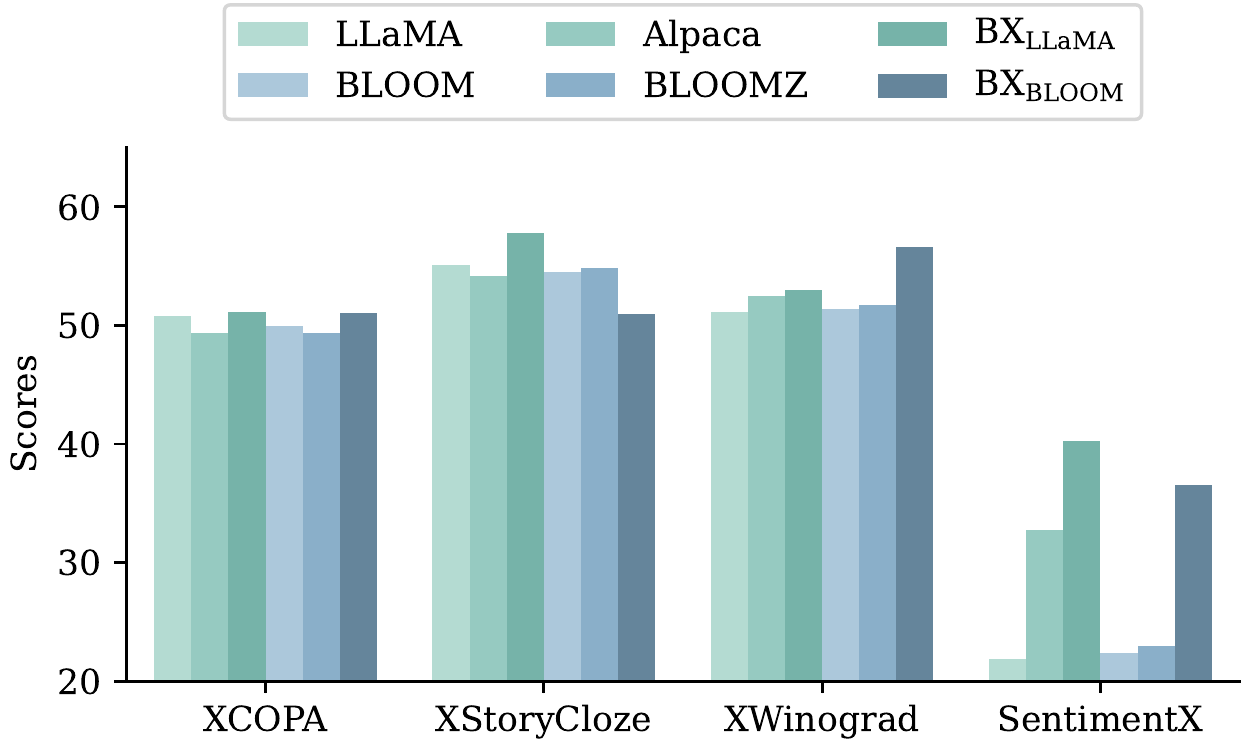}
	\caption{The average performance of 7B models on unseen languages (i.e.~languages that are not used in pre-training the base model). }
	\label{fig:experiment_downstream}
\end{figure}

\begin{table}[t]
    \small
    \centering
    \resizebox{\linewidth}{!}{
        \begin{tabular}{lcccc}
            \toprule
            {Tasks} & {\bxllama}  & BM$_\text{LLaMA}$ & {\bxbloom} & BM$_\text{BLOOM}$ \\
            %\cmidrule{2-3} \cmidrule{5-6}
            %& Multi & Mono && Multi & Mono \\
            \midrule
            XCOPA & 52.2 & \textbf{52.7} &   56.0 & \textbf{56.6} \\
            XStoryCloze & 59.6 & \textbf{60.5} &   59.1 & \textbf{60.7} \\
            XWinograd & 61.6 & \textbf{64.2} &   61.7 & \textbf{64.1} \\
            SentimentX & 44.1 & \textbf{44.2} &   31.3 & \textbf{41.6} \\
            \midrule
            Average & 54.4 & \textbf{55.4} &   52.0 & \textbf{55.8} \\
            \bottomrule
        \end{tabular}
        }
    \caption{Zero-shot performance of multilingual BX and monolingual BM models with 7B parameters. We report averaged accuracy for \texttt{XCOPA}, \texttt{XStoryCloze}, and \texttt{XWinograd}, and averaged weighted F1-macro scores for \texttt{SentimentX}.}
    \label{tab:mono_multi}
\end{table}

\paragraph{Performance on Unseen Languages}
In \figref{fig:experiment_downstream}, we present the average performance of the 7B models over languages that the base models were not exposed to in pre-training. For \texttt{XCOPA}, \texttt{XStoryCloze}, \texttt{XWinograd}, and \texttt{SentimentX}, the LLaMA model is not exposed to 10, 8, 2, and 5 languages, resp., while the BLOOM model is not exposed to 7, 2, 2, and 4 languages, respectively. 
We observe that our proposed models improve on the zero-shot performance of the base models across all tasks, and also surpass the performance of existing instruction-tuned models, with the exception of BLOOM over \texttt{XStoryCloze}. A notable improvement can be seen in the \texttt{SentimentX} dataset, implying that our models are more suited to non-English instructions and non-English sentiment labels.
% When comparing it with \tabref{tab:experiment_downstream}, where BLOOMZ outperforms \bxbloom in the \texttt{SentimentX} task, we conjecture that the \texttt{xP3} dataset contains specific data related to this task. In contrast, our instruction dataset consists of more general instructions, which helps the model generalize better to unseen languages.

\paragraph{Monolingual vs.\ Multilingual Fine-tuning} 
For each of the 52 languages in Section~\ref{sec:data}, 
we compared the performance of monolingual BM models against the multilingual BX models. To ensure a fair benchmark, we exclude unseen languages in calculating the average score. Table~\ref{tab:mono_multi} presents the average performance for each dataset, revealing that the monolingual BM models consistently outperform the multilingual model for both LLaMA and BLOOM. 
Particularly notable improvements are observed for \texttt{XWinograd} and \texttt{SentimentX}. For example, the monolingual BM$_\text{BLOOM}$ achieves an impressive overall increase of $+10.3$ compared to the multilingual model for \texttt{SentimentX}.

\begin{table}[t]
\small
\centering
\setlength{\tabcolsep}{4pt}
\begin{tabular}{lccc}
\toprule
Models & Natural & Social & Others \\ 
\midrule
LLaMA (13B) & 30.09 & 32.77 & 31.11 \\
Alpaca (13B) & 28.19 & 32.99 & 30.36 \\
\bxllama (13B) & \textbf{33.58} & \textbf{39.15} & \textbf{39.71} \\
\bottomrule
\end{tabular}
\caption{
    Performance breakdown by subject type in \texttt{EXAMS}. ``Natural'' and ``Social'' denote natural science and social science, respectively.
}
\label{tab:exam_subjects}
\end{table}

\subsection{Knowledge-intensive Task}\label{sec:exam_results}

The last column of \tabref{tab:experiment_downstream}  shows the results on \texttt{EXAMS}, averaged across languages. 
We find that the \bxllama models (7B and 13B) outperform their corresponding base models, while BLOOMZ outperforms our \bxbloom.
We observe that multilingual instruction tuning seems to be more promising on larger models, as seen in \bxllama (13B) improving substantially over LLaMA by 5.5\% on average, while the margin for \bxllama (7B) is only 0.9\%. It is noteworthy that \bxllama (13B) also outperforms LLaMA (30B) on the \texttt{EXAMS} benchmark in Table \ref{tab:exam} in \appref{app:all_res}, underlining the effectiveness of multilingual instruction tuning.

% \paragraph{Result Breakdown by Subjects}
The \texttt{EXAMS} dataset comprises a range of subject areas, such as natural science and social science. 
We present a breakdown of the results across subject areas for the 13B models in Table \ref{tab:exam_subjects}. 
It is evident that there are substantial performance improvements over the social sciences and other subject areas during fine-tuning, but comparatively lesser gains for natural science. 
This could be attributed to our dataset containing fewer instructions and questions related to natural sciences, or the inherent difficulty of learning natural science concepts or reasoning abilities through instruction fine-tuning.

\begin{figure}[t]
    \centering
    \footnotesize
    \begin{tcolorbox}[colback=white,arc=0mm,left=0mm,right=0mm,top=1mm,bottom=1mm,colframe=color3,width=\linewidth]
\textcolor{teal}{You are a helpful and precise assistant for checking the quality of the answer.} \\

<question> \\
Comment les obstacles linguistiques et culturels ... \\
</question> \\
<answer1> \\
Les obstacles linguistiques peuvent avoir un impact ... \\
</answer1> \\
<answer2> \\
The linguistic and cultural obstacles ... \\
</answer2> \\

\textcolor{teal}{Suppose the user only speaks the language of the question, please evaluate both answers with your justification having less three sentences, and provide a score ranging from 0 to 10 after your justifications. When evaluating the answers, you should consider the helpfulness, relevance, accuracy, level of details of the answers. The score for answer 1 should be wrapped by <score1> and </score1>, and the score for answer 2 should be wrapped by <score2> and </score2>.}
    \end{tcolorbox}
    \caption{Template for GPT-4 evaluation. The colored parts are
      general prompts that are used for all instances.}
    \label{fig:gpt4_prompt_example}

\end{figure}

\section{Evaluation on Open-ended Questions}
\label{sec:general_eval}

As LLMs continue to develop, existing NLP benchmarks may not be up to the task of evaluating model capabilities. To address this, we use GPT-4 \cite{gpt4} as an evaluator to compare model outputs, supplemented by human evaluations.

We adopt a challenging set of 80 questions covering 8 categories from \citet{vicuna2023} for open-ended question evaluation. These questions are translated into 51 languages, and we use different models to generate responses (see \appref{app:output_examples} for examples). 
Following \citet{vicuna2023}, we provide two answers from different models in a single prompt, and ask GPT-4 to rate the answers over a scale of 0 to 10 from various aspects including helpfulness, relevance, accuracy, and the level of detail (see \figref{fig:gpt4_prompt_example} for an example prompt for GPT-4 evaluation). 
To ensure fairness, we interchange the order of the provided answers, and assign scores twice for each question.
We exclude vanilla BLOOM and LLaMA from open-ended question evaluation, and instead compare \bxbloom against BLOOMZ, \bxllama against Alpaca, and \bxbloom against \bxllama, given the superiority of instruction-tuned models in previous studies \citep{vicuna2023,bloomz}.
We select 5 questions from each category, resulting in 40 questions per language. Given cost restrictions and availability of human annotators, we conducted GPT-4 evaluation over 12 languages and human evaluation over 6 languages.

\begin{figure}[t]
	\centering
	\includegraphics[width=\linewidth]{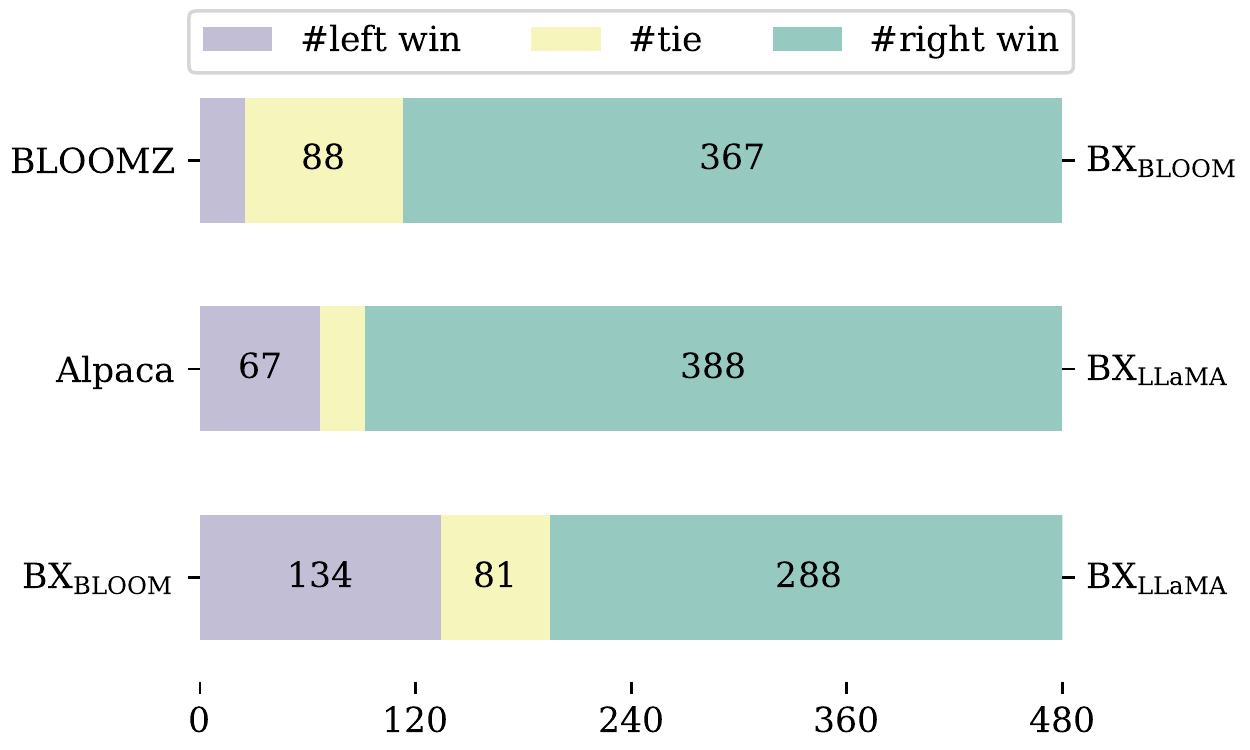}
	\caption{Overall comparison of GPT-4 evaluation. }
	\label{fig:gpt4_main}
\end{figure}

\subsection{GPT-4 Evaluation}
\figref{fig:gpt4_main} shows the results of the three model pairs, clearly indicate that GPT-4 has a preference for \bxllama over Alpaca and similarly favors \bxbloom over BLOOMZ. Regarding the comparison between the two BX models, \bxllama performs better overall.

\begin{figure}[t]
	\centering
	\includegraphics[width=\linewidth]{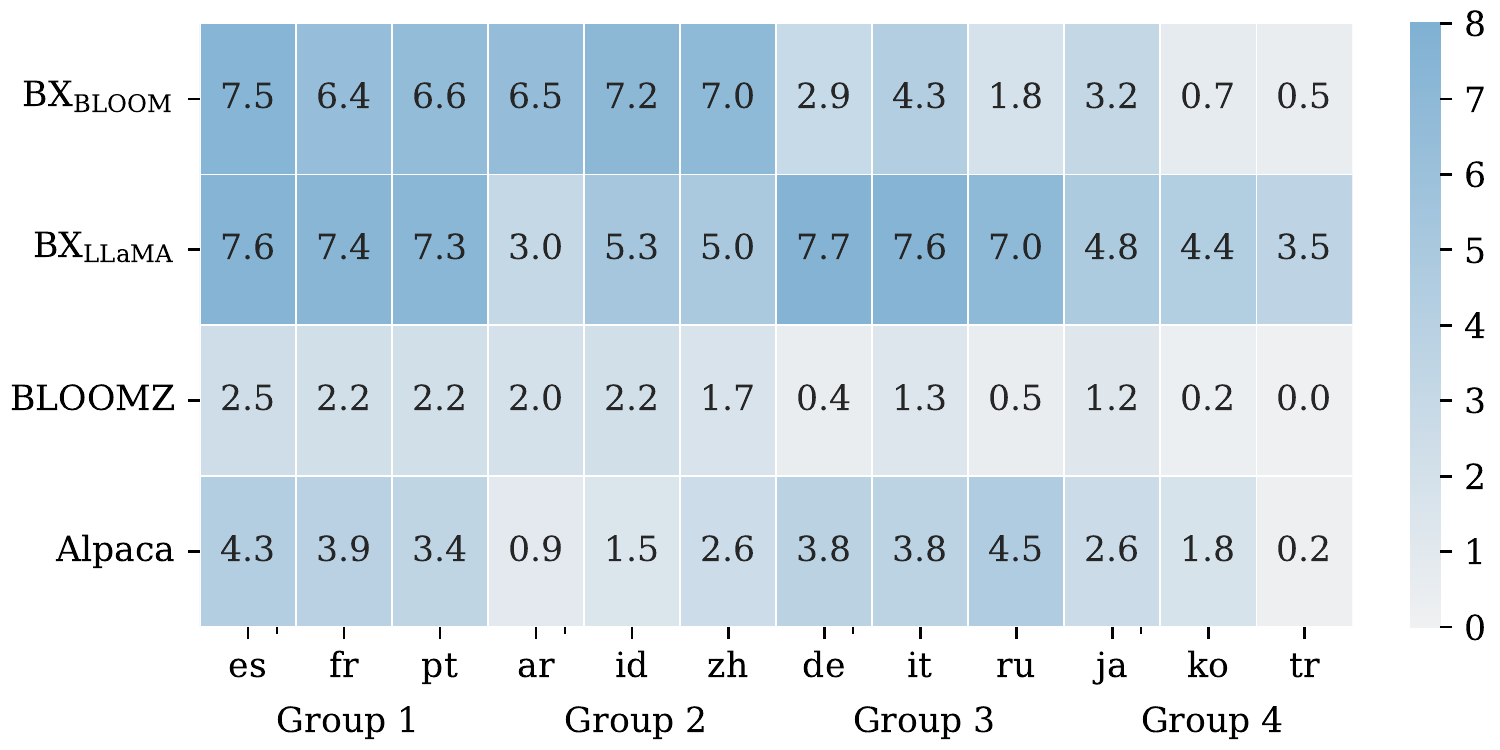}
	\caption{GPT-4 evaluation by language. We categorize languages into four groups based on whether a language is seen during model pre-training, and select 3 languages from each group.
 \texttt{Group 1}: languages seen by both BLOOM and LLaMA; \texttt{group} 2: seen by BLOOM only;
 \texttt{group 3}: seen by LLaMA only; \texttt{group 4}: not seen by either BLOOM or LLaMA.}
	\label{fig:gpt4_lang}
\end{figure}

\begin{figure}[t]
	\centering
	\includegraphics[width=\linewidth]{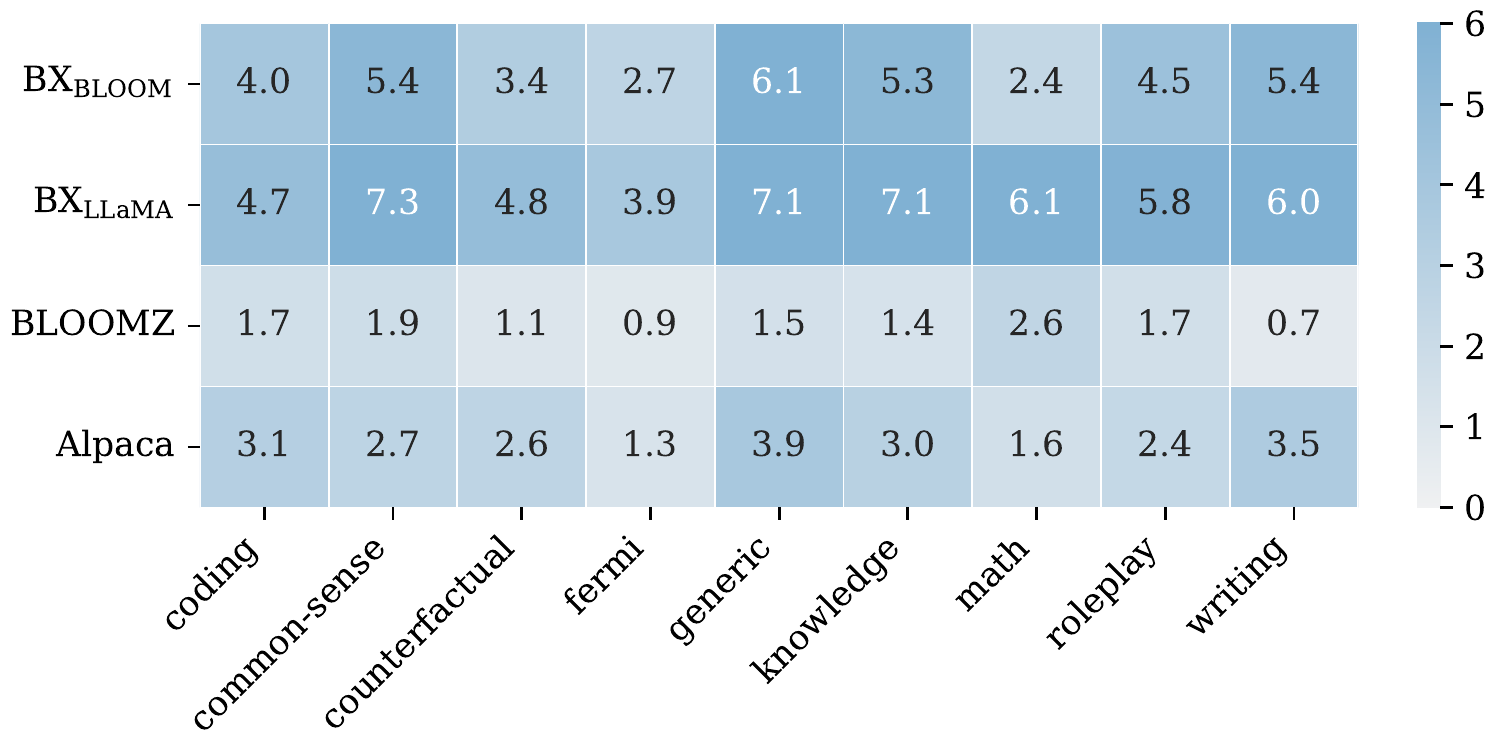}
	\caption{GPT-4 evaluation by question types.}
	\label{fig:gpt4_qtype}
\end{figure}

Since GPT-4 assigns a quantitative score to each response on a scale of 0--10, we calculate the average score for each model from all comparison pairs and present a breakdown of results separately for each language group (see Figure \ref{fig:gpt4_lang}) and question type (see Figure \ref{fig:gpt4_qtype}).

\paragraph{Language Group}
Analyzing the results based by language group (see \figref{fig:gpt4_lang}), we can make several observations. 
First, multilingual pre-training plays a critical role for multilingual instruction-following models. In groups 1 and 3, \bxllama outperforms \bxbloom, while in group 2, \bxbloom performs substantially better. This difference can be attributed to variations in language coverage during pre-training, as both models are fine-tuned on the same dataset.
Second, multilingual instruction-tuning is critical. \bxllama, fine-tuned on our multilingual dataset, outperforms Alpaca, which is only fine-tuned on English instructions, across all evaluated languages.
From group 4, we observe that if a language is not included in pre-training, multilingual instruction-tuning alone is insufficient to achieve strong performance.
Additionally, both \bxbloom and BLOOMZ are initialized by BLOOM but fine-tuned on different instruction datasets. BLOOMZ is fine-tuned on xP3, a multilingual instruction dataset based on hand-written templates and downstream NLP tasks. In this free generation evaluation, \bxbloom performs much better than BLOOMZ, highlighting the limitations of human-written instructions in terms of diversity.
Overall, multilinguality in both pre-training and instruction-tuning is vital for the effectiveness of multilingual instruction-following models. These findings reinforce our contributions in this work. 

\paragraph{Question Type}
When considering different question types (see \figref{fig:gpt4_qtype}), the Bactrian-X models consistently outperform all base models.
A noteworthy observation is that ``fermi'' and ``math'' questions, which require strong reasoning capabilities, prove to be challenging for all multilingual LLMs. This observation underlines the fact that numerical reasoning task in a multilingual setup remains an under-explored area, requiring further research.

\subsection{Human Evaluation}

We conducted human evaluation of the outputs of four models (LLaMA, \bxllama, BLOOMZ, and \bxbloom) for the six languages as before, namely three high-resource languages --- Arabic (\texttt{ar}), Indonesian (\texttt{id}), Chinese (\texttt{zh}) --- and three low-resource languages --- Burmese (\texttt{my}), Tamil (\texttt{ta}), and Tagalog (\texttt{tl}). Native-speaker annotators were asked to rank the outputs of these models based on their overall quality, from 1 (best) to 4 (worst). Prior to annotation, models are shuffled and their identities are not visible to the annotators.

The average Spearman rank correlation between annotators is $\rho = 0.78$ across languages, indicating high inter-annotator agreement. 

The human evaluation results, averaged across languages and models, are presented in Table~\ref{tab:human_eval}. 
Overall, we observe that our models \bxbloom and \bxllama are better than their instruction-tuned counterparts BLOOMZ and Alpaca, once again emphasizing the effectiveness of our multilingual dataset and language adaptation technique. In particular, \bxbloom achieves superior performance for \texttt{ar}, \texttt{id}, \texttt{zh}, and \texttt{ta}, which are languages included in the pre-training of BLOOM. On the other hand, \bxllama performs the best over \texttt{my} and \texttt{tl}, which are unseen languages for both base models.

\begin{table}[t]
\small
\centering
\begin{tabular}{@{}lrrrrrr@{}}
\toprule
\multirow{2}{*}{{Model}} &  \multicolumn{6}{c}{Language} \\
\cmidrule{2-7}
& \texttt{ar} & \texttt{id} & \texttt{zh} & \texttt{my} & \texttt{ta} & \texttt{tl} \\
\midrule
Alpaca & 16.7 & 11.7 & 7.1 & 59.6 & 2.1 & 51.7 \\
\bxllama & 69.2 & 71.3 & 78.3 & \textbf{92.1} & 46.7 & \textbf{81.7} \\
BLOOMZ & 27.5 & 37.1 & 30.0 & 20.0 & 67.5 & 7.5 \\
\bxbloom & \textbf{86.7} & \textbf{80.0} & \textbf{84.6} & 28.3 & \textbf{83.8} & 59.2 \\
\bottomrule
\end{tabular}
\caption{
   The results of human evaluation for model responses across six languages.
   We map the ranks 1, 2, 3, and 4 into scores 100, 66, 33, and 0, respectively, and then average the two annotator scores.
   Higher is better. Note that the number here represents a relative ranking score; therefore, a high score does not indicate a perfect model.
}
\label{tab:human_eval}
\end{table}

% \begin{table}[t]
% \small
% \centering
% \begin{tabular}{@{}lrrrrrr@{}}
% \toprule
% \multirow{2}{*}{{Model}} &  \multicolumn{6}{c}{Language} \\
% \cmidrule{2-7}
% & \texttt{ar} & \texttt{id} & \texttt{zh} & \texttt{my} & \texttt{ta} & \texttt{tl} \\
% \midrule
% Alpaca & 16.7 & 11.7 & 7.1 & 59.6 & 2.1 & 51.7 \\
% \bxllama & 69.2 & 71.3 & 78.3 & \textbf{92.1} & 46.7 & \textbf{81.7} \\
% BLOOMZ & 27.5 & 37.1 & 30.0 & 20.0 & 67.5 & 7.5 \\
% \bxbloom & \textbf{86.7} & \textbf{80.0} & \textbf{84.6} & 28.3 & \textbf{83.8} & 59.2 \\
% \bottomrule
% \end{tabular}
% \caption{
%    The results of human evaluation for model responses across six languages.
%    We average the two annotator ranking.
%    Lower is better.
% }
% \label{tab:human_eval}
% \end{table}

%%% Local Variables:
%%% mode: latex
%%% TeX-master: "emnlp2023"
%%% End:
 
\section{Conclusion}

% In this paper, we introduce Bactrian-X, a comprehensive multilingual parallel dataset consisting of 3.4 million instruction-response pairs across 52 languages. This dataset is openly available to the research community. We create a series of adapters trained on our datasets, which are lightweight components that can be seamlessly integrated with foundational models. With less than 1\% of the parameters, these adapters can be easily replaced and used as plug-ins for different languages or language groups. Through our experiments, we demonstrate that the models fine-tuned on the Bactrian-X datasets surpass their corresponding vanilla models and exhibit superior performance compared to other instruction-tuned model versions in various multilingual downstream tasks. We hope that the availability of this dataset can accelerate the development of large language models in multilingual scenarios, facilitating advancements in natural language processing across diverse languages.
%We plan to release larger models fine-tuned on our dataset, employing full precision (float32), in future endeavors.

In this paper, we have introduced \datasetname, a comprehensive multilingual parallel dataset comprising 3.4 million instruction--response pairs across 52 languages.
To enhance the multilingual capabilities of base LLMs, we also introduced a collection of lightweight adapters trained on \datasetname. 
Experiments on various multilingual NLP tasks demonstrate that models fine-tuned on the Bactrian-X dataset outperform both their corresponding vanilla models and also models fine-tuned on other monolingual/multilingual instruction datasets.
By making our dataset and models available, we hope to expedite the advancement of LLMs for multilingual purposes, promoting progress in natural language processing across a broader set of languages.

%%% Local Variables:
%%% mode: latex
%%% TeX-master: "emnlp2023"
%%% End:

\section*{Limitations}

Our work is subject to several limitations that should be addressed in future research:
(1) Our focus was limited to 7B and 13B models, without exploring scaling rules or other base models such as mT5~\citep{xue-etal-2021-mt5}. Further investigation into different model variations could provide valuable insights.
(2) In our experiments, the maximum sequence length for multilingual models was set to 768 sub-word units. This smaller context size, compared to models with lengths of 1024 or 2048, may restrict the model's ability to effectively leverage long-range context. Additionally, certain languages that were not well supported by the model tokenizers could face challenges with such a small context size.
(3) We did not thoroughly investigate the presence of hallucination, toxicity, and fairness in our models or the base models due to the unavailability of an appropriate evaluation suite. Nonetheless, it is important to acknowledge that our models, as well as the base models, are likely to be susceptible to these concerns. Future research should address these issues to ensure responsible and unbiased model behavior.
We acknowledge these limitations and propose that future work should focus on addressing them to advance the utility and deployment-safety of the models.

\section*{Ethical Considerations}
While our instruction-tuning datasets and models offer several advantages, it is essential to recognize their limitations. Despite efforts made by ChatGPT to alleviate ethical concerns, it is still possible for the model to generate responses that are discriminatory, biased, or contain false information, particularly in multilingual settings. Hence, our models, when fine-tuned on the dataset, may inadvertently learn or propagate these problematic patterns.

To address these concerns and minimize potential harm, we are dedicated to mitigating the risks associated with the use of our models in future research. We strongly advocate for the responsible use of our models to prevent any unintended negative consequences.

%%% Local Variables:
%%% mode: latex
%%% TeX-master: "emnlp2023"
%%% End:

% \input{8_ethics}
% \input{9_conclusion}

\bibliography{anthology,custom}

\begin{thebibliography}{41}
\expandafter\ifx\csname natexlab\endcsname\relax\def\natexlab#1{#1}\fi

\bibitem[{Alturayeif et~al.(2022)Alturayeif, Luqman, and
  Ahmed}]{alturayeif-etal-2022-mawqif}
Nora~Saleh Alturayeif, Hamzah~Abdullah Luqman, and Moataz Aly~Kamaleldin Ahmed.
  2022.
\newblock \href {https://aclanthology.org/2022.wanlp-1.16} {Mawqif: A
  multi-label {A}rabic dataset for target-specific stance detection}.
\newblock In \emph{Proceedings of the The Seventh Arabic Natural Language
  Processing Workshop (WANLP)}, pages 174--184, Abu Dhabi, United Arab Emirates
  (Hybrid). Association for Computational Linguistics.

\bibitem[{Ben~Zaken et~al.(2022)Ben~Zaken, Goldberg, and
  Ravfogel}]{ben-zaken-etal-2022-bitfit}
Elad Ben~Zaken, Yoav Goldberg, and Shauli Ravfogel. 2022.
\newblock \href {https://doi.org/10.18653/v1/2022.acl-short.1} {{B}it{F}it:
  Simple parameter-efficient fine-tuning for transformer-based masked
  language-models}.
\newblock In \emph{Proceedings of the 60th Annual Meeting of the Association
  for Computational Linguistics (Volume 2: Short Papers)}, pages 1--9, Dublin,
  Ireland. Association for Computational Linguistics.

\bibitem[{Brown et~al.(2020)Brown, Mann, Ryder, Subbiah, Kaplan, Dhariwal,
  Neelakantan, Shyam, Sastry, Askell, Agarwal, Herbert-Voss, Krueger, Henighan,
  Child, Ramesh, Ziegler, Wu, Winter, Hesse, Chen, Sigler, Litwin, Gray, Chess,
  Clark, Berner, McCandlish, Radford, Sutskever, and
  Amodei}]{NEURIPS2020_1457c0d6}
Tom Brown, Benjamin Mann, Nick Ryder, Melanie Subbiah, Jared~D Kaplan, Prafulla
  Dhariwal, Arvind Neelakantan, Pranav Shyam, Girish Sastry, Amanda Askell,
  Sandhini Agarwal, Ariel Herbert-Voss, Gretchen Krueger, Tom Henighan, Rewon
  Child, Aditya Ramesh, Daniel Ziegler, Jeffrey Wu, Clemens Winter, Chris
  Hesse, Mark Chen, Eric Sigler, Mateusz Litwin, Scott Gray, Benjamin Chess,
  Jack Clark, Christopher Berner, Sam McCandlish, Alec Radford, Ilya Sutskever,
  and Dario Amodei. 2020.
\newblock \href
  {https://proceedings.neurips.cc/paper_files/paper/2020/file/1457c0d6bfcb4967418bfb8ac142f64a-Paper.pdf}
  {Language models are few-shot learners}.
\newblock In \emph{Advances in Neural Information Processing Systems},
  volume~33, pages 1877--1901. Curran Associates, Inc.

\bibitem[{Chiang et~al.(2023)Chiang, Li, Lin, Sheng, Wu, Zhang, Zheng, Zhuang,
  Zhuang, Gonzalez, Stoica, and Xing}]{vicuna2023}
Wei-Lin Chiang, Zhuohan Li, Zi~Lin, Ying Sheng, Zhanghao Wu, Hao Zhang, Lianmin
  Zheng, Siyuan Zhuang, Yonghao Zhuang, Joseph~E. Gonzalez, Ion Stoica, and
  Eric~P. Xing. 2023.
\newblock \href {https://vicuna.lmsys.org} {Vicuna: An open-source chatbot
  impressing gpt-4 with 90\%* chatgpt quality}.

\bibitem[{Chowdhery et~al.(2022)Chowdhery, Narang, Devlin, Bosma, Mishra,
  Roberts, Barham, Chung, Sutton, Gehrmann, Schuh, Shi, Tsvyashchenko, Maynez,
  Rao, Barnes, Tay, Shazeer, Prabhakaran, Reif, Du, Hutchinson, Pope, Bradbury,
  Austin, Isard, Gur{-}Ari, Yin, Duke, Levskaya, Ghemawat, Dev, Michalewski,
  Garcia, Misra, Robinson, Fedus, Zhou, Ippolito, Luan, Lim, Zoph, Spiridonov,
  Sepassi, Dohan, Agrawal, Omernick, Dai, Pillai, Pellat, Lewkowycz, Moreira,
  Child, Polozov, Lee, Zhou, Wang, Saeta, Diaz, Firat, Catasta, Wei,
  Meier{-}Hellstern, Eck, Dean, Petrov, and
  Fiedel}]{DBLP:journals/corr/abs-2204-02311}
Aakanksha Chowdhery, Sharan Narang, Jacob Devlin, Maarten Bosma, Gaurav Mishra,
  Adam Roberts, Paul Barham, Hyung~Won Chung, Charles Sutton, Sebastian
  Gehrmann, Parker Schuh, Kensen Shi, Sasha Tsvyashchenko, Joshua Maynez,
  Abhishek Rao, Parker Barnes, Yi~Tay, Noam Shazeer, Vinodkumar Prabhakaran,
  Emily Reif, Nan Du, Ben Hutchinson, Reiner Pope, James Bradbury, Jacob
  Austin, Michael Isard, Guy Gur{-}Ari, Pengcheng Yin, Toju Duke, Anselm
  Levskaya, Sanjay Ghemawat, Sunipa Dev, Henryk Michalewski, Xavier Garcia,
  Vedant Misra, Kevin Robinson, Liam Fedus, Denny Zhou, Daphne Ippolito, David
  Luan, Hyeontaek Lim, Barret Zoph, Alexander Spiridonov, Ryan Sepassi, David
  Dohan, Shivani Agrawal, Mark Omernick, Andrew~M. Dai,
  Thanumalayan~Sankaranarayana Pillai, Marie Pellat, Aitor Lewkowycz, Erica
  Moreira, Rewon Child, Oleksandr Polozov, Katherine Lee, Zongwei Zhou, Xuezhi
  Wang, Brennan Saeta, Mark Diaz, Orhan Firat, Michele Catasta, Jason Wei,
  Kathy Meier{-}Hellstern, Douglas Eck, Jeff Dean, Slav Petrov, and Noah
  Fiedel. 2022.
\newblock \href {https://doi.org/10.48550/arXiv.2204.02311} {Palm: Scaling
  language modeling with pathways}.
\newblock \emph{CoRR}, abs/2204.02311.

\bibitem[{Chung et~al.(2022)Chung, Hou, Longpre, Zoph, Tay, Fedus, Li, Wang,
  Dehghani, Brahma, Webson, Gu, Dai, Suzgun, Chen, Chowdhery, Narang, Mishra,
  Yu, Zhao, Huang, Dai, Yu, Petrov, Chi, Dean, Devlin, Roberts, Zhou, Le, and
  Wei}]{DBLP:journals/corr/abs-2210-11416}
Hyung~Won Chung, Le~Hou, Shayne Longpre, Barret Zoph, Yi~Tay, William Fedus,
  Eric Li, Xuezhi Wang, Mostafa Dehghani, Siddhartha Brahma, Albert Webson,
  Shixiang~Shane Gu, Zhuyun Dai, Mirac Suzgun, Xinyun Chen, Aakanksha
  Chowdhery, Sharan Narang, Gaurav Mishra, Adams Yu, Vincent~Y. Zhao, Yanping
  Huang, Andrew~M. Dai, Hongkun Yu, Slav Petrov, Ed~H. Chi, Jeff Dean, Jacob
  Devlin, Adam Roberts, Denny Zhou, Quoc~V. Le, and Jason Wei. 2022.
\newblock \href {https://doi.org/10.48550/arXiv.2210.11416} {Scaling
  instruction-finetuned language models}.
\newblock \emph{CoRR}, abs/2210.11416.

\bibitem[{Conover et~al.(2023)Conover, Hayes, Mathur, Meng, Xie, Wan, Shah,
  Ghodsi, Wendell, Zaharia, and Xin}]{dolly}
Mike Conover, Matt Hayes, Ankit Mathur, Xiangrui Meng, Jianwei Xie, Jun Wan,
  Sam Shah, Ali Ghodsi, Patrick Wendell, Matei Zaharia, and Reynold Xin. 2023.
\newblock Free dolly: Introducing the world's first truly open
  instruction-tuned llm.
\newblock \url{https://www.databricks.com}.

\bibitem[{Guo et~al.(2021)Guo, Rush, and Kim}]{guo-etal-2021-parameter}
Demi Guo, Alexander Rush, and Yoon Kim. 2021.
\newblock \href {https://doi.org/10.18653/v1/2021.acl-long.378}
  {Parameter-efficient transfer learning with diff pruning}.
\newblock In \emph{Proceedings of the 59th Annual Meeting of the Association
  for Computational Linguistics and the 11th International Joint Conference on
  Natural Language Processing (Volume 1: Long Papers)}, pages 4884--4896,
  Online. Association for Computational Linguistics.

\bibitem[{Hardalov et~al.(2020)Hardalov, Mihaylov, Zlatkova, Dinkov, Koychev,
  and Nakov}]{exams}
Momchil Hardalov, Todor Mihaylov, Dimitrina Zlatkova, Yoan Dinkov, Ivan
  Koychev, and Preslav Nakov. 2020.
\newblock \href {https://doi.org/10.18653/v1/2020.emnlp-main.438} {{EXAMS:} {A}
  multi-subject high school examinations dataset for cross-lingual and
  multilingual question answering}.
\newblock In \emph{Proceedings of the 2020 Conference on Empirical Methods in
  Natural Language Processing, {EMNLP} 2020, Online, November 16-20, 2020},
  pages 5427--5444. Association for Computational Linguistics.

\bibitem[{Hayashibe(2020)}]{hayashibe-2020-japanese}
Yuta Hayashibe. 2020.
\newblock \href {https://aclanthology.org/2020.lrec-1.843} {{J}apanese
  realistic textual entailment corpus}.
\newblock In \emph{Proceedings of the Twelfth Language Resources and Evaluation
  Conference}, pages 6827--6834, Marseille, France. European Language Resources
  Association.

\bibitem[{Hoffmann et~al.(2022)Hoffmann, Borgeaud, Mensch, Buchatskaya, Cai,
  Rutherford, de~Las~Casas, Hendricks, Welbl, Clark, Hennigan, Noland,
  Millican, van~den Driessche, Damoc, Guy, Osindero, Simonyan, Elsen, Rae,
  Vinyals, and Sifre}]{DBLP:journals/corr/abs-2203-15556}
Jordan Hoffmann, Sebastian Borgeaud, Arthur Mensch, Elena Buchatskaya, Trevor
  Cai, Eliza Rutherford, Diego de~Las~Casas, Lisa~Anne Hendricks, Johannes
  Welbl, Aidan Clark, Tom Hennigan, Eric Noland, Katie Millican, George van~den
  Driessche, Bogdan Damoc, Aurelia Guy, Simon Osindero, Karen Simonyan, Erich
  Elsen, Jack~W. Rae, Oriol Vinyals, and Laurent Sifre. 2022.
\newblock \href {https://doi.org/10.48550/arXiv.2203.15556} {Training
  compute-optimal large language models}.
\newblock \emph{CoRR}, abs/2203.15556.

\bibitem[{Houlsby et~al.(2019)Houlsby, Giurgiu, Jastrzebski, Morrone,
  De~Laroussilhe, Gesmundo, Attariyan, and Gelly}]{pmlr-v97-houlsby19a}
Neil Houlsby, Andrei Giurgiu, Stanislaw Jastrzebski, Bruna Morrone, Quentin
  De~Laroussilhe, Andrea Gesmundo, Mona Attariyan, and Sylvain Gelly. 2019.
\newblock \href {https://proceedings.mlr.press/v97/houlsby19a.html}
  {Parameter-efficient transfer learning for {NLP}}.
\newblock In \emph{Proceedings of the 36th International Conference on Machine
  Learning}, volume~97 of \emph{Proceedings of Machine Learning Research},
  pages 2790--2799. PMLR.

\bibitem[{Hu et~al.(2022)Hu, yelong shen, Wallis, Allen-Zhu, Li, Wang, Wang,
  and Chen}]{hu2022lora}
Edward~J Hu, yelong shen, Phillip Wallis, Zeyuan Allen-Zhu, Yuanzhi Li, Shean
  Wang, Lu~Wang, and Weizhu Chen. 2022.
\newblock \href {https://openreview.net/forum?id=nZeVKeeFYf9} {Lo{RA}: Low-rank
  adaptation of large language models}.
\newblock In \emph{International Conference on Learning Representations}.

\bibitem[{Koto et~al.(2020)Koto, Rahimi, Lau, and
  Baldwin}]{koto-etal-2020-indolem}
Fajri Koto, Afshin Rahimi, Jey~Han Lau, and Timothy Baldwin. 2020.
\newblock \href {https://doi.org/10.18653/v1/2020.coling-main.66} {{I}ndo{LEM}
  and {I}ndo{BERT}: A benchmark dataset and pre-trained language model for
  {I}ndonesian {NLP}}.
\newblock In \emph{Proceedings of the 28th International Conference on
  Computational Linguistics}, pages 757--770, Barcelona, Spain (Online).
  International Committee on Computational Linguistics.

\bibitem[{Lester et~al.(2021)Lester, Al-Rfou, and
  Constant}]{lester-etal-2021-power}
Brian Lester, Rami Al-Rfou, and Noah Constant. 2021.
\newblock \href {https://doi.org/10.18653/v1/2021.emnlp-main.243} {The power of
  scale for parameter-efficient prompt tuning}.
\newblock In \emph{Proceedings of the 2021 Conference on Empirical Methods in
  Natural Language Processing}, pages 3045--3059, Online and Punta Cana,
  Dominican Republic. Association for Computational Linguistics.

\bibitem[{Lin et~al.(2022)Lin, Mihaylov, Artetxe, Wang, Chen, Simig, Ott,
  Goyal, Bhosale, Du, Pasunuru, Shleifer, Koura, Chaudhary, O{'}Horo, Wang,
  Zettlemoyer, Kozareva, Diab, Stoyanov, and Li}]{lin-etal-2022-shot}
Xi~Victoria Lin, Todor Mihaylov, Mikel Artetxe, Tianlu Wang, Shuohui Chen,
  Daniel Simig, Myle Ott, Naman Goyal, Shruti Bhosale, Jingfei Du, Ramakanth
  Pasunuru, Sam Shleifer, Punit~Singh Koura, Vishrav Chaudhary, Brian O{'}Horo,
  Jeff Wang, Luke Zettlemoyer, Zornitsa Kozareva, Mona Diab, Veselin Stoyanov,
  and Xian Li. 2022.
\newblock \href {https://aclanthology.org/2022.emnlp-main.616} {Few-shot
  learning with multilingual generative language models}.
\newblock In \emph{Proceedings of the 2022 Conference on Empirical Methods in
  Natural Language Processing}, pages 9019--9052, Abu Dhabi, United Arab
  Emirates. Association for Computational Linguistics.

\bibitem[{Mangrulkar et~al.(2022)Mangrulkar, Gugger, Debut, Belkada, and
  Paul}]{peft}
Sourab Mangrulkar, Sylvain Gugger, Lysandre Debut, Younes Belkada, and Sayak
  Paul. 2022.
\newblock Peft: State-of-the-art parameter-efficient fine-tuning methods.
\newblock \url{https://github.com/huggingface/peft}.

\bibitem[{Mostafazadeh et~al.(2016)Mostafazadeh, Chambers, He, Parikh, Batra,
  Vanderwende, Kohli, and Allen}]{mostafazadeh-etal-2016-corpus}
Nasrin Mostafazadeh, Nathanael Chambers, Xiaodong He, Devi Parikh, Dhruv Batra,
  Lucy Vanderwende, Pushmeet Kohli, and James Allen. 2016.
\newblock \href {https://doi.org/10.18653/v1/N16-1098} {A corpus and cloze
  evaluation for deeper understanding of commonsense stories}.
\newblock In \emph{Proceedings of the 2016 Conference of the North {A}merican
  Chapter of the Association for Computational Linguistics: Human Language
  Technologies}, pages 839--849, San Diego, California. Association for
  Computational Linguistics.

\bibitem[{Muennighoff et~al.(2022)Muennighoff, Wang, Sutawika, Roberts,
  Biderman, Scao, Bari, Shen, Yong, Schoelkopf, Tang, Radev, Aji, Almubarak,
  Albanie, Alyafeai, Webson, Raff, and Raffel}]{bloomz}
Niklas Muennighoff, Thomas Wang, Lintang Sutawika, Adam Roberts, Stella
  Biderman, Teven~Le Scao, M.~Saiful Bari, Sheng Shen, Zheng~Xin Yong, Hailey
  Schoelkopf, Xiangru Tang, Dragomir Radev, Alham~Fikri Aji, Khalid Almubarak,
  Samuel Albanie, Zaid Alyafeai, Albert Webson, Edward Raff, and Colin Raffel.
  2022.
\newblock \href {https://doi.org/10.48550/arXiv.2211.01786} {Crosslingual
  generalization through multitask finetuning}.
\newblock \emph{CoRR}, abs/2211.01786.

\bibitem[{Muhammad et~al.(2023)Muhammad, Abdulmumin, Yimam, Adelani, Ahmad,
  Ousidhoum, Ayele, Mohammad, Beloucif, and Ruder}]{muhammadSemEval2023}
Shamsuddeen~Hassan Muhammad, Idris Abdulmumin, Seid~Muhie Yimam, David~Ifeoluwa
  Adelani, Ibrahim~Sa'id Ahmad, Nedjma Ousidhoum, Abinew~Ali Ayele, Saif~M.
  Mohammad, Meriem Beloucif, and Sebastian Ruder. 2023.
\newblock {SemEval-2023 Task 12: Sentiment Analysis for African Languages
  (AfriSenti-SemEval)}.
\newblock In \emph{Proceedings of the 17th {{International Workshop}} on
  {{Semantic Evaluation}} ({{SemEval-2023}})}. {Association for Computational
  Linguistics}.

\bibitem[{OpenAI(2023)}]{gpt4}
OpenAI. 2023.
\newblock \href {https://doi.org/10.48550/arXiv.2303.08774} {{GPT-4} technical
  report}.
\newblock \emph{CoRR}, abs/2303.08774.

\bibitem[{Ouyang et~al.(2022)Ouyang, Wu, Jiang, Almeida, Wainwright, Mishkin,
  Zhang, Agarwal, Slama, Gray, Schulman, Hilton, Kelton, Miller, Simens,
  Askell, Welinder, Christiano, Leike, and Lowe}]{ouyang2022training}
Long Ouyang, Jeffrey Wu, Xu~Jiang, Diogo Almeida, Carroll Wainwright, Pamela
  Mishkin, Chong Zhang, Sandhini Agarwal, Katarina Slama, Alex Gray, John
  Schulman, Jacob Hilton, Fraser Kelton, Luke Miller, Maddie Simens, Amanda
  Askell, Peter Welinder, Paul Christiano, Jan Leike, and Ryan Lowe. 2022.
\newblock \href {https://openreview.net/forum?id=TG8KACxEON} {Training language
  models to follow instructions with human feedback}.
\newblock In \emph{Advances in Neural Information Processing Systems}.

\bibitem[{Papineni et~al.(2002)Papineni, Roukos, Ward, and
  Zhu}]{papineni-etal-2002-bleu}
Kishore Papineni, Salim Roukos, Todd Ward, and Wei-Jing Zhu. 2002.
\newblock \href {https://doi.org/10.3115/1073083.1073135} {{B}leu: a method for
  automatic evaluation of machine translation}.
\newblock In \emph{Proceedings of the 40th Annual Meeting of the Association
  for Computational Linguistics}, pages 311--318, Philadelphia, Pennsylvania,
  USA. Association for Computational Linguistics.

\bibitem[{Ponti et~al.(2020)Ponti, {s}, Majewska, Liu, Vuli'{c}, and
  Korhonen}]{ponti2020xcopa}
Edoardo~M. Ponti, Goran~Glava {s}, Olga Majewska, Qianchu Liu, Ivan Vuli'{c},
  and Anna Korhonen. 2020.
\newblock \href {https://ducdauge.github.io/files/xcopa.pdf} {{XCOPA: A}
  multilingual dataset for causal commonsense reasoning}.
\newblock \emph{arXiv preprint}.

\bibitem[{Popovi{\'c}(2017)}]{popovic-2017-chrf}
Maja Popovi{\'c}. 2017.
\newblock \href {https://doi.org/10.18653/v1/W17-4770} {chr{F}++: words helping
  character n-grams}.
\newblock In \emph{Proceedings of the Second Conference on Machine
  Translation}, pages 612--618, Copenhagen, Denmark. Association for
  Computational Linguistics.

\bibitem[{Post(2018)}]{post-2018-call}
Matt Post. 2018.
\newblock \href {https://doi.org/10.18653/v1/W18-6319} {A call for clarity in
  reporting {BLEU} scores}.
\newblock In \emph{Proceedings of the Third Conference on Machine Translation:
  Research Papers}, pages 186--191, Brussels, Belgium. Association for
  Computational Linguistics.

\bibitem[{Rei et~al.(2020)Rei, Stewart, Farinha, and
  Lavie}]{rei-etal-2020-comet}
Ricardo Rei, Craig Stewart, Ana~C Farinha, and Alon Lavie. 2020.
\newblock \href {https://doi.org/10.18653/v1/2020.emnlp-main.213} {{COMET}: A
  neural framework for {MT} evaluation}.
\newblock In \emph{Proceedings of the 2020 Conference on Empirical Methods in
  Natural Language Processing (EMNLP)}, pages 2685--2702, Online. Association
  for Computational Linguistics.

\bibitem[{Sanh et~al.(2022)Sanh, Webson, Raffel, Bach, Sutawika, Alyafeai,
  Chaffin, Stiegler, Raja, Dey, Bari, Xu, Thakker, Sharma, Szczechla, Kim,
  Chhablani, Nayak, Datta, Chang, Jiang, Wang, Manica, Shen, Yong, Pandey,
  Bawden, Wang, Neeraj, Rozen, Sharma, Santilli, F{\'{e}}vry, Fries, Teehan,
  Scao, Biderman, Gao, Wolf, and Rush}]{DBLP:conf/iclr/SanhWRBSACSRDBX22}
Victor Sanh, Albert Webson, Colin Raffel, Stephen~H. Bach, Lintang Sutawika,
  Zaid Alyafeai, Antoine Chaffin, Arnaud Stiegler, Arun Raja, Manan Dey,
  M~Saiful Bari, Canwen Xu, Urmish Thakker, Shanya~Sharma Sharma, Eliza
  Szczechla, Taewoon Kim, Gunjan Chhablani, Nihal~V. Nayak, Debajyoti Datta,
  Jonathan Chang, Mike~Tian{-}Jian Jiang, Han Wang, Matteo Manica, Sheng Shen,
  Zheng~Xin Yong, Harshit Pandey, Rachel Bawden, Thomas Wang, Trishala Neeraj,
  Jos Rozen, Abheesht Sharma, Andrea Santilli, Thibault F{\'{e}}vry, Jason~Alan
  Fries, Ryan Teehan, Teven~Le Scao, Stella Biderman, Leo Gao, Thomas Wolf, and
  Alexander~M. Rush. 2022.
\newblock \href {https://openreview.net/forum?id=9Vrb9D0WI4} {Multitask
  prompted training enables zero-shot task generalization}.
\newblock In \emph{The Tenth International Conference on Learning
  Representations, {ICLR} 2022, Virtual Event, April 25-29, 2022}.
  OpenReview.net.

\bibitem[{Scao et~al.(2022)Scao, Fan, Akiki, Pavlick, Ilic, Hesslow,
  Castagn{\'{e}}, Luccioni, Yvon, Gall{\'{e}}, Tow, Rush, Biderman, Webson,
  Ammanamanchi, Wang, Sagot, Muennighoff, del Moral, Ruwase, Bawden, Bekman,
  McMillan{-}Major, Beltagy, Nguyen, Saulnier, Tan, Suarez, Sanh,
  Lauren{\c{c}}on, Jernite, Launay, Mitchell, Raffel, Gokaslan, Simhi, Soroa,
  Aji, Alfassy, Rogers, Nitzav, Xu, Mou, Emezue, Klamm, Leong, van Strien,
  Adelani, and et~al.}]{bloom}
Teven~Le Scao, Angela Fan, Christopher Akiki, Ellie Pavlick, Suzana Ilic,
  Daniel Hesslow, Roman Castagn{\'{e}}, Alexandra~Sasha Luccioni,
  Fran{\c{c}}ois Yvon, Matthias Gall{\'{e}}, Jonathan Tow, Alexander~M. Rush,
  Stella Biderman, Albert Webson, Pawan~Sasanka Ammanamanchi, Thomas Wang,
  Beno{\^{\i}}t Sagot, Niklas Muennighoff, Albert~Villanova del Moral, Olatunji
  Ruwase, Rachel Bawden, Stas Bekman, Angelina McMillan{-}Major, Iz~Beltagy,
  Huu Nguyen, Lucile Saulnier, Samson Tan, Pedro~Ortiz Suarez, Victor Sanh,
  Hugo Lauren{\c{c}}on, Yacine Jernite, Julien Launay, Margaret Mitchell, Colin
  Raffel, Aaron Gokaslan, Adi Simhi, Aitor Soroa, Alham~Fikri Aji, Amit
  Alfassy, Anna Rogers, Ariel~Kreisberg Nitzav, Canwen Xu, Chenghao Mou, Chris
  Emezue, Christopher Klamm, Colin Leong, Daniel van Strien, David~Ifeoluwa
  Adelani, and et~al. 2022.
\newblock \href {https://doi.org/10.48550/arXiv.2211.05100} {{BLOOM:} {A}
  176b-parameter open-access multilingual language model}.
\newblock \emph{CoRR}, abs/2211.05100.

\bibitem[{Tang et~al.(2020)Tang, Tran, Li, Chen, Goyal, Chaudhary, Gu, and
  Fan}]{tang2020multilingual}
Yuqing Tang, Chau Tran, Xian Li, Peng-Jen Chen, Naman Goyal, Vishrav Chaudhary,
  Jiatao Gu, and Angela Fan. 2020.
\newblock Multilingual translation with extensible multilingual pretraining and
  finetuning.
\newblock \emph{arXiv preprint arXiv:2008.00401}.

\bibitem[{Taori et~al.(2023)Taori, Gulrajani, Zhang, Dubois, Li, Guestrin,
  Liang, and Hashimoto}]{alpaca}
Rohan Taori, Ishaan Gulrajani, Tianyi Zhang, Yann Dubois, Xuechen Li, Carlos
  Guestrin, Percy Liang, and Tatsunori~B. Hashimoto. 2023.
\newblock Stanford alpaca: An instruction-following llama model.
\newblock \url{https://github.com/tatsu-lab/stanford_alpaca}.

\bibitem[{Tikhonov and Ryabinin(2021)}]{tikhonov-ryabinin-2021-heads}
Alexey Tikhonov and Max Ryabinin. 2021.
\newblock \href {https://doi.org/10.18653/v1/2021.findings-acl.310} {{I}t{'}s
  {A}ll in the {H}eads: {U}sing {A}ttention {H}eads as a {B}aseline for
  {C}ross-{L}ingual {T}ransfer in {C}ommonsense {R}easoning}.
\newblock In \emph{Findings of the Association for Computational Linguistics:
  ACL-IJCNLP 2021}, pages 3534--3546, Online. Association for Computational
  Linguistics.

\bibitem[{Touvron et~al.(2023)Touvron, Lavril, Izacard, Martinet, Lachaux,
  Lacroix, Rozi{\`{e}}re, Goyal, Hambro, Azhar, Rodriguez, Joulin, Grave, and
  Lample}]{llama}
Hugo Touvron, Thibaut Lavril, Gautier Izacard, Xavier Martinet, Marie{-}Anne
  Lachaux, Timoth{\'{e}}e Lacroix, Baptiste Rozi{\`{e}}re, Naman Goyal, Eric
  Hambro, Faisal Azhar, Aur{\'{e}}lien Rodriguez, Armand Joulin, Edouard Grave,
  and Guillaume Lample. 2023.
\newblock \href {https://doi.org/10.48550/arXiv.2302.13971} {Llama: Open and
  efficient foundation language models}.
\newblock \emph{CoRR}, abs/2302.13971.

\bibitem[{Vaswani et~al.(2017)Vaswani, Shazeer, Parmar, Uszkoreit, Jones,
  Gomez, Kaiser, and Polosukhin}]{vaswani2017attention}
Ashish Vaswani, Noam Shazeer, Niki Parmar, Jakob Uszkoreit, Llion Jones,
  Aidan~N Gomez, {\L}ukasz Kaiser, and Illia Polosukhin. 2017.
\newblock Attention is all you need.
\newblock \emph{Advances in neural information processing systems}, 30.

\bibitem[{Wang et~al.(2022)Wang, Kordi, Mishra, Liu, Smith, Khashabi, and
  Hajishirzi}]{self-instruct}
Yizhong Wang, Yeganeh Kordi, Swaroop Mishra, Alisa Liu, Noah~A. Smith, Daniel
  Khashabi, and Hannaneh Hajishirzi. 2022.
\newblock \href {https://doi.org/10.48550/arXiv.2212.10560} {Self-instruct:
  Aligning language model with self generated instructions}.
\newblock \emph{CoRR}, abs/2212.10560.

\bibitem[{Wei et~al.(2022)Wei, Bosma, Zhao, Guu, Yu, Lester, Du, Dai, and
  Le}]{wei2022finetuned}
Jason Wei, Maarten Bosma, Vincent Zhao, Kelvin Guu, Adams~Wei Yu, Brian Lester,
  Nan Du, Andrew~M. Dai, and Quoc~V Le. 2022.
\newblock \href {https://openreview.net/forum?id=gEZrGCozdqR} {Finetuned
  language models are zero-shot learners}.
\newblock In \emph{International Conference on Learning Representations}.

\bibitem[{Winata et~al.(2023)Winata, Aji, Cahyawijaya, Mahendra, Koto,
  Romadhony, Kurniawan, Moeljadi, Prasojo, Fung, Baldwin, Lau, Sennrich, and
  Ruder}]{winata-etal-2023-nusax}
Genta~Indra Winata, Alham~Fikri Aji, Samuel Cahyawijaya, Rahmad Mahendra, Fajri
  Koto, Ade Romadhony, Kemal Kurniawan, David Moeljadi, Radityo~Eko Prasojo,
  Pascale Fung, Timothy Baldwin, Jey~Han Lau, Rico Sennrich, and Sebastian
  Ruder. 2023.
\newblock \href {https://aclanthology.org/2023.eacl-main.57} {{N}usa{X}:
  Multilingual parallel sentiment dataset for 10 {I}ndonesian local languages}.
\newblock In \emph{Proceedings of the 17th Conference of the European Chapter
  of the Association for Computational Linguistics}, pages 815--834, Dubrovnik,
  Croatia. Association for Computational Linguistics.

\bibitem[{Wolf et~al.(2020)Wolf, Debut, Sanh, Chaumond, Delangue, Moi, Cistac,
  Rault, Louf, Funtowicz, Davison, Shleifer, von Platen, Ma, Jernite, Plu, Xu,
  Le~Scao, Gugger, Drame, Lhoest, and Rush}]{wolf-etal-2020-transformers}
Thomas Wolf, Lysandre Debut, Victor Sanh, Julien Chaumond, Clement Delangue,
  Anthony Moi, Pierric Cistac, Tim Rault, Remi Louf, Morgan Funtowicz, Joe
  Davison, Sam Shleifer, Patrick von Platen, Clara Ma, Yacine Jernite, Julien
  Plu, Canwen Xu, Teven Le~Scao, Sylvain Gugger, Mariama Drame, Quentin Lhoest,
  and Alexander Rush. 2020.
\newblock \href {https://doi.org/10.18653/v1/2020.emnlp-demos.6} {Transformers:
  State-of-the-art natural language processing}.
\newblock In \emph{Proceedings of the 2020 Conference on Empirical Methods in
  Natural Language Processing: System Demonstrations}, pages 38--45, Online.
  Association for Computational Linguistics.

\bibitem[{Wu et~al.(2023)Wu, Waheed, Zhang, Abdul-Mageed, and Aji}]{lamini-lm}
Minghao Wu, Abdul Waheed, Chiyu Zhang, Muhammad Abdul-Mageed, and Alham~Fikri
  Aji. 2023.
\newblock \href {http://arxiv.org/abs/2304.14402} {Lamini-lm: A diverse herd of
  distilled models from large-scale instructions}.
\newblock \emph{CoRR}, abs/2304.14402.

\bibitem[{Xue et~al.(2021)Xue, Constant, Roberts, Kale, Al-Rfou, Siddhant,
  Barua, and Raffel}]{xue-etal-2021-mt5}
Linting Xue, Noah Constant, Adam Roberts, Mihir Kale, Rami Al-Rfou, Aditya
  Siddhant, Aditya Barua, and Colin Raffel. 2021.
\newblock \href {https://doi.org/10.18653/v1/2021.naacl-main.41} {m{T}5: A
  massively multilingual pre-trained text-to-text transformer}.
\newblock In \emph{Proceedings of the 2021 Conference of the North American
  Chapter of the Association for Computational Linguistics: Human Language
  Technologies}, pages 483--498, Online. Association for Computational
  Linguistics.

\bibitem[{Zeng et~al.(2023)Zeng, Liu, Du, Wang, Lai, Ding, Yang, Xu, Zheng,
  Xia, Tam, Ma, Xue, Zhai, Chen, Liu, Zhang, Dong, and Tang}]{zeng2023glmb}
Aohan Zeng, Xiao Liu, Zhengxiao Du, Zihan Wang, Hanyu Lai, Ming Ding, Zhuoyi
  Yang, Yifan Xu, Wendi Zheng, Xiao Xia, Weng~Lam Tam, Zixuan Ma, Yufei Xue,
  Jidong Zhai, Wenguang Chen, Zhiyuan Liu, Peng Zhang, Yuxiao Dong, and Jie
  Tang. 2023.
\newblock \href {https://openreview.net/forum?id=-Aw0rrrPUF} {{GLM}-130b: An
  open bilingual pre-trained model}.
\newblock In \emph{The Eleventh International Conference on Learning
  Representations}.

\end{thebibliography}
\bibliographystyle{acl_natbib}

\clearpage
\appendix

\section{Annotation guidelines for response quality checking}\label{app:quality_guideline}
We asked the human experts to rate fluency and informativeness separately, following the guidelines in \figref{fig:info_guideline} and \figref{fig:fluent_guideline} separately.

\begin{figure}[h]
	\small
	\begin{tcolorbox}[colframe=white, left=3mm, right=1mm]
		Read the input, and judge/mark the output: \\
  
\textbf{Rating-A}: The output is valid, factually correct, and satisfying. \\
\textbf{Rating-B}: The output is acceptable with minor errors. \\
\textbf{Rating-C}: The output is relevant but has significant errors.  \\
\textbf{Rating-D}: The output is completely bad. \\
\textbf{Rating-E}: I don't know. \\
	\end{tcolorbox}
	\caption{Annotation guidelines for response informativeness.}
	\label{fig:info_guideline}
\end{figure}

\begin{figure}[h]
	\small
	\begin{tcolorbox}[colframe=white, left=3mm, right=1mm]
Read the input, and judge/mark the output: \\

\textbf{Rating-A}: High fluency, like a native speaker! \\
\textbf{Rating-B}: Moderate fluency (generally coherent, with minor errors). \\
\textbf{Rating-C}: Low fluency (noticable errors). \\
\textbf{Rating-D}: Not fluent at all, or the output is in a different language. \\
	\end{tcolorbox}
	\caption{Annotation guidelines for response fluency.}
	\label{fig:fluent_guideline}
\end{figure}

\section{Base models}\label{app:models}

\begin{itemize}
    \item LLaMA \citep{llama}: a series of base models proposed by Meta, encompassing a parameter range of 7B to 65B. The models were primarily trained on English, but include around 4.5\% of text from 20 different languages in the training data, enabling some level of support for multilingual tasks.
    \item Alpaca \citep{alpaca}: a fine-tuned variant of the LLaMA model on 52K English instruction-following data instances generated through self-instruct techniques \citep{self-instruct}. In initial human evaluation, the 7B Alpaca model was observed to attain similar behavior to the text-davinci-003 model (130B) on the self-instruct instruction-following evaluation suite \citep{self-instruct}.
    \item BLOOM \cite{bloom}: a collection of pretraiend multilingual language models created by BigScience, trained on the ROOTS corpus, which encompasses data from 46 languages. 
    \item BLOOMZ \citep{bloomz}: derived from BLOOM and fine-tuned using the crosslingual task mixture (xP3) dataset, and capable of zero-shot instruction-following in dozens of languages. 
\end{itemize}

\section{Hyperparameters for Bactrian-X models}\label{sec:hyper_parameters}
The hyperparameters for the \datasetname models are shown in Table \ref{tab:hyper_parameters}. It is important to note that during the fine-tuning process, the instructions are masked, and the loss is computed only for the responses. This approach effectively prevents the models from learning ``translationese'' and allows it to focus on distilling ChatGPT's responses.
\begin{table}[ht]
    \small
    \centering
    \begin{tabular}{lll}
        \toprule
        Hyper-parameter	& Multi & Mono \\
        \midrule
        batch size & 128 & 128 \\
        steps & 100k & 5k \\
        learning rate & 3e-4  & 3e-4 \\
        max seq length & 768 & 1024 \\
        lora r & 64 & 16 \\
        lora alpha & 16 & 16 \\ 
        lora dropout & 0.05 & 0.05 \\
        \bottomrule
    \end{tabular}
    \caption{Hyperparameters for multilingual and monolingual model training.}
    \label{tab:hyper_parameters}
\end{table}

\section{Complete results for the multilingual benchmark}\label{app:all_res}
\begin{table*}[ht]
\tiny
\centering
\begin{tabular}{@{}lcccccccccccc@{}}
\toprule
Models & et & ht & id & it & qu & sw & ta & th & tr & vi & zh & Avg \\ \midrule
LLaMA-7B & 49.80 & 50.00 & 51.80 & 52.40 & 51.60 & 49.20 & 45.60 & 52.60 & 49.80 & 49.80 & 49.80 & 50.22 \\
Alpaca-LoRA-7B & 48.20 & 50.40 & 53.00 & 59.00 & 50.20 & 49.20 & 44.40 & 48.20 & 49.60 & 47.80 & 52.80 & 50.25 \\
\bxllama-7B & 52.40 & 48.40 & 52.80 & 59.20 & 51.60 & 52.60 & 45.40 & 53.00 & 50.40 & 49.20 & 54.40 & \textbf{51.76} \\ \midrule
LLama-13B & 51.00 & 50.40 & 52.20 & 55.60 & 50.40 & 49.00 & 46.40 & 51.80 & 50.60 & 51.00 & 53.00 & 51.04 \\
Alpaca-LoRA-13B & 47.40 & 52.80 & 57.80 & 73.20 & 50.40 & 52.60 & 47.80 & 52.60 & 52.60 & 51.60 & 64.20 & \textbf{54.82} \\
\bxllama-13B & 53.80 & 49.20 & 56.20 & 64.80 & 49.40 & 52.60 & 45.60 & 52.00 & 51.20 & 53.20 & 58.00 & 53.27 \\ \midrule
BLOOM-7B & 48.00 & 46.00 & 59.20 & 48.60 & 52.00 & 49.60 & 44.80 & 51.40 & 52.40 & 61.60 & 57.80 & 51.95 \\
BLOOMZ-7B  & 49.20 & 43.40 & 59.40 & 49.40 & 52.00 & 51.60 & 45.60 & 50.00 & 52.00 & 61.40 & 59.40 & 52.13 \\
\bxbloom-7B & 50.80 & 47.80 & 65.40 & 54.40 & 50.60 & 52.60 & 46.00 & 53.80 & 52.20 & 63.20 & 65.80 & \textbf{54.78} \\
\bottomrule
\end{tabular}
\caption{
    Accuracy of zero-shot performance over \texttt{XCOPA}.
}
\label{tab:xcopa}
\end{table*}

\begin{table*}[ht]
\tiny
\centering
\begin{tabular}{@{}lccccccccccc@{}}
\toprule
Models & ar & es & eu & hi & id & my & ru & sw & te & zh & Avg \\ \midrule
LLaMA-7B & 53.47 & 62.08 & 52.02 & 55.72 & 57.58 & 55.13 & 62.54 & 55.33 & 58.70 & 57.71 & 57.03 \\
Alpaca-LoRA-7B & 51.26 & 64.88 & 51.92 & 54.23 & 57.08 & 54.17 & 61.84 & 55.15 & 57.93 & 59.06 & 56.75 \\
\bxllama-7B & 54.67 & 67.57 & 52.28 & 56.32 & 59.56 & 57.78 & 65.85 & 57.31 & 57.71 & 60.03 & \textbf{58.91} \\ \midrule
LLama-13B & 53.41 & 65.59 & 53.74 & 54.40 & 59.17 & 54.40 & 64.26 & 55.79 & 57.51 & 60.56 & 57.88 \\
Alpaca-LoRA-13B & 54.40 & 71.81 & 53.08 & 55.33 & 57.58 & 52.88 & 71.48 & 55.00 & 57.18 & 61.55 & 59.03 \\
\bxllama-13B & 57.11 & 76.70 & 53.28 & 58.84 & 62.41 & 57.45 & 72.87 & 60.16 & 56.85 & 65.59 & \textbf{62.12} \\ \midrule
BLOOM-7B & 56.65 & 59.36 & 54.14 & 51.16 & 61.09 & 54.53 & 56.59 & 55.66 & 52.48 & 63.67 & 56.53 \\
BLOOMZ-7B  & 60.29 & 64.79 & 55.13 & 51.69 & 62.28 & 54.86 & 56.98 & 56.92 & 52.08 & 65.52 & 58.05 \\
\bxbloom-7B & 58.97 & 68.83 & 53.74 & 50.76 & 68.03 & 50.96 & 57.05 & 56.92 & 52.02 & 68.30 & \textbf{58.56} \\
\bottomrule
\end{tabular}
\caption{
    Accuracy of zero-shot performance over \texttt{XStoryCloze}.
}
\label{tab:xcloze}
\end{table*}

\begin{table*}[ht]
\tiny
\centering
\begin{tabular}{@{}lccccccc@{}}
\toprule
Models & en & fr & jp & pt & zh & ru & Avg \\ \midrule
LLaMA-7B & 63.66 & 56.63 & 51.09 & 56.65 & 59.72 & 60.00 & 57.96 \\ 
Alpaca-LoRA-7B & 65.63 & 56.63 & 52.45 & 55.51 & 57.54 & 58.41 & 57.70 \\ 
\bxllama-7B & 68.13 & 60.24 & 52.97 & 58.17 & 61.11 & 60.32 & \textbf{60.16} \\ \midrule
LLama-13B & 54.00 & 51.81 & 51.00 & 52.00 & 56.00 & 53.00 & 52.97 \\
Alpaca-LoRA-13B & 55.00 & 50.60 & 47.00 & 50.00 & 61.00 & 50.00 & 52.27 \\
\bxllama-13B & 72.34 & 61.45 & 54.54 & 66.54 & 62.90 & 64.13 & \textbf{63.65} \\ \midrule
BLOOM-7B & 60.65 & 59.04 & 51.41 & 57.79 & 65.28 & 53.65 & 57.97 \\
BLOOMZ-7B  & 65.63 & 62.65 & 51.72 & 58.17 & 67.86 & 54.29 & 60.05 \\ 
\bxbloom-7B & 66.28 & 55.42 & 56.62 & 63.12 & 70.83 & 52.70 & \textbf{60.83} \\

\bottomrule
\end{tabular}
\caption{
    Accuracy of zero-shot performance over \texttt{XWinograd}.
}
\label{tab:xwino}
\end{table*}

\begin{table*}[ht]
\tiny
\centering
\begin{tabular}{@{}lccccccccc@{}}
\toprule
Models & ar & es & jp & ru & id & jav & sun & sw & Avg \\ \midrule
LLaMA-7B & 26.79 & 29.27 & 4.58 & 46.49 & 35.54 & 34.49 & 26.47 & 44.23 & 30.98 \\
Alpaca-LoRA-7B & 34.56 & 56.05 & 43.28 & 12.73 & 35.95 & 23.88 & 31.00 & 42.82 & 35.03 \\
\bxllama-7B & 31.19 & 54.90 & 51.44 & 56.29 & 34.09 & 30.12 & 39.20 & 43.99 & \textbf{42.65} \\ \midrule
LLama-13B & 36.41 & 31.32 & 46.25 & 3.46 & 35.47 & 33.39 & 37.00 & 44.90 & 33.52 \\
Alpaca-LoRA-13B & 51.16 & 52.30 & 30.94 & 10.85 & 55.08 & 40.51 & 30.21 & 15.30 & 35.79 \\
\bxllama-13B & 36.42 & 66.82 & 54.90 & 63.13 & 55.00 & 40.73 & 40.65 & 44.50 & \textbf{50.27} \\ \midrule
BLOOM-7B & 23.39 & 31.04 & 6.25 & 68.36 & 21.63 & 23.19 & 37.67 & 3.53 & 26.88 \\
BLOOMZ-7B  & 48.68 & 40.22 & 3.49 & 68.58 & 40.53 & 27.08 & 38.24 & 34.61 & \textbf{37.68} \\
\bxbloom-7B & 14.57 & 36.88 & 46.34 & 68.19 & 23.27 & 25.27 & 38.11 & 13.62 & 33.28 \\ \bottomrule
\end{tabular}
\caption{
    Macro-F1 scores of zero-shot performance over \texttt{SentimentX}.
}
\label{tab:sentimentx}
\end{table*}

\begin{table*}[ht]
\tiny
\centering
\setlength{\tabcolsep}{4pt}
\begin{tabular}{@{}lccccccccccccccccc@{}}
\toprule
 & ar & bg & de & es & fr & hr & hu & it & lt & mk & pl & pt & sq & sr & tr & vi & Avg \\ \midrule
BLOOM-7B & 24.0 & 24.4 & 24.5 & 21.1 & 32.8 & 24.1 & 27.2 & 24.3 & 24.6 & 24.7 & 25.0 & 24.2 & 23.7 & 25.5 & 24.7 & 26.1 & 25.1 \\
BLOOMZ-7B & 38.4 & 27.7 & 30.2 & 28.9 & 40.6 & 28.4 & 26.3 & 37.9 & 24.8 & 27.5 & 27.4 & 43.3 & 28.8 & 26.4 & 27.7 & 35.3 & 31.2 \\
\bxbloom-7B & 27.2 & 26.0 & 23.6 & 36.8 & 32.0 & 26.3 & 26.3 & 24.4 & 24.3 & 23.4 & 24.7 & 26.3 & 23.3 & 26.0 & 23.2 & 25.3 & 26.2 \\ \midrule
LLaMA-7B & 24.2 & 28.3 & 33.0 & 28.9 & 29.7 & 29.6 & 28.7 & 31.8 & 27.2 & 28.9 & 27.3 & 25.3 & 29.0 & 27.8 & 26.7 & 24.8 & 28.2 \\
Alpaca-7B & 25.4 & 30.8 & 31.6 & 26.3 & 29.7 & 32.0 & 27.7 & 32.0 & 25.6 & 30.5 & 28.8 & 31.0 & 29.4 & 28.7 & 26.1 & 25.6 & 28.8 \\ 
\bxllama-7B & 25.8 & 30.5 & 32.1 & 31.6 & 35.9 & 30.1 & 25.9 & 31.9 & 25.0 & 31.3 & 28.7 & 29.4 & 26.2 & 27.0 & 28.6 & 26.3 & 29.1 \\ \midrule
LLaMA-13B & 23.5 & 34.3 & 25.5 & 31.6 & 33.6 & 36.5 & 27.3 & 34.1 & 25.6 & 35.3 & 28.9 & 31.6 & 30.1 & 33.0 & 29.3 & 26.4 & 30.4 \\
Alpaca-13B & 26.3 & 33.5 & 29.7 & 28.9 & 35.9 & 33.3 & 28.7 & 33.6 & 30.0 & 34.3 & 29.4 & 31.3 & 27.8 & 32.0 & 28.0 & 24.6 & 30.5 \\
\bxllama-13B & 29.9 & 41.9 & 40.1 & 31.6 & 33.6 & 41.8 & 35.4 & 41.1 & 28.8 & 44.8 & 36.9 & 34.8 & 31.2 & 36.3 & 35.0 & 28.2 & 35.7 \\ \midrule
LLaMA-30B & 22.6 & 33.6 & 33.5 & 39.5 & 28.9 & 32.9 & 27.2 & 35.2 & 29.8 & 32.5 & 31.2 & 31.4 & 26.9 & 30.9 & 28.3 & 25.8 & 30.6 \\
LLaMA-65B & 32.4 & 52.1 & 47.6 & 44.7 & 50.0 & 53.4 & 40.2 & 51.2 & 34.2 & 57.2 & 44.6 & 49.5 & 40.5 & 47.4 & 42.4 & 27.7 & 44.7 \\
\bottomrule
\end{tabular}
\caption{
     Accuracy of zero-shot performance over \texttt{EXAMS}.
}
\label{tab:exam}
\end{table*}

\begin{table*}[ht]
\small
\centering
\begin{tabular}{@{}lccc@{}}
\toprule
Language & Test size & Seen by BLOOM & Seen by LLaMA \\ \midrule
et & 500 & no & no \\
ht & 500 & no & no \\
id & 500 & yes & no \\
it & 500 & no & yes \\
qu & 500 & no & no \\
sw & 500 & yes & no \\
ta & 500 & yes & no \\
th & 500 & no & no \\
tr & 500 & no & no \\ 
vi & 500 & yes & no \\
zh & 500 & yes & no \\
\bottomrule
\end{tabular}
\caption{
    \texttt{XCOPA} data statistics.
}
\label{tab:xcopa_stat}
\end{table*}

\begin{table*}[ht]
\small
\centering
\begin{tabular}{@{}lccc@{}}
\toprule
Language & Test size & Seen by BLOOM & Seen by LLaMA \\ \midrule
ar & 1511 & yes & no \\
es & 1511 & yes & yes \\
eu & 1511 & yes & no \\
hi & 1511 & yes & no \\
id & 1511 & yes & no \\
my & 1511 & no & no \\ 
ru & 1511 & no & yes \\
sw & 1511 & yes & no \\
te & 1511 & yes & no \\ 
zh & 1511 & yes & no \\
\bottomrule
\end{tabular}
\caption{
    \texttt{XStoryCloze} data statistics.
}
\label{tab:xcloze_stat}
\end{table*}

\begin{table*}[ht]
\small
\centering
\begin{tabular}{@{}lccc@{}}
\toprule
Language & Test size & Seen by BLOOM & Seen by LLaMA \\ \midrule
en & 2325 & yes & yes \\ 
fr & 83 & yes & yes \\
jp & 959 & no & no \\
pt & 263 & yes & yes \\ 
zh & 504 & yes & no \\
ru & 315 & no & yes \\ 
\bottomrule
\end{tabular}
\caption{
    \texttt{XWinograd} data statistics.
}
\label{tab:xwino_stat}
\end{table*}

\begin{table*}[ht]
\small
\centering
\begin{tabular}{@{}lccc@{}}
\toprule
Language & Test size & Seen by BLOOM & Seen by LLaMA \\ \midrule
ar & 619 & yes & no \\
es & 1000 & yes & yes \\ 
jp & 553 & no & no \\
ru & 1000 & no & yes \\ 
id & 400 & yes & no \\ 
jav & 400 & no & no \\
sun & 400 & no & no \\
\bottomrule
\end{tabular}
\caption{
    \texttt{SentimentX} data statistics. For Spanish (es) and Russian
    (ru), we randomly sample 1000 test instances.
}
\label{tab:sentimentx_stat}
\end{table*}

We present the full zero-shot results for the multilingual benchmark in \tabref{tab:xcopa} (\texttt{XCOPA}), \tabref{tab:xcloze} (\texttt{XStoryCloze}), \tabref{tab:xwino} (\texttt{XWinograd}), and \tabref{tab:sentimentx} (\texttt{SentimentX}). Please refer to \tabref{tab:xcopa_stat}, \tabref{tab:xcloze_stat}, \tabref{tab:xwino_stat}, \tabref{tab:sentimentx_stat} for details of the data distributions used for evaluation.

\section{Model output examples in 9 different languages}\label{app:output_examples}
\begin{figure*}[h]
    \centering
    \includegraphics[width=.9\textwidth]{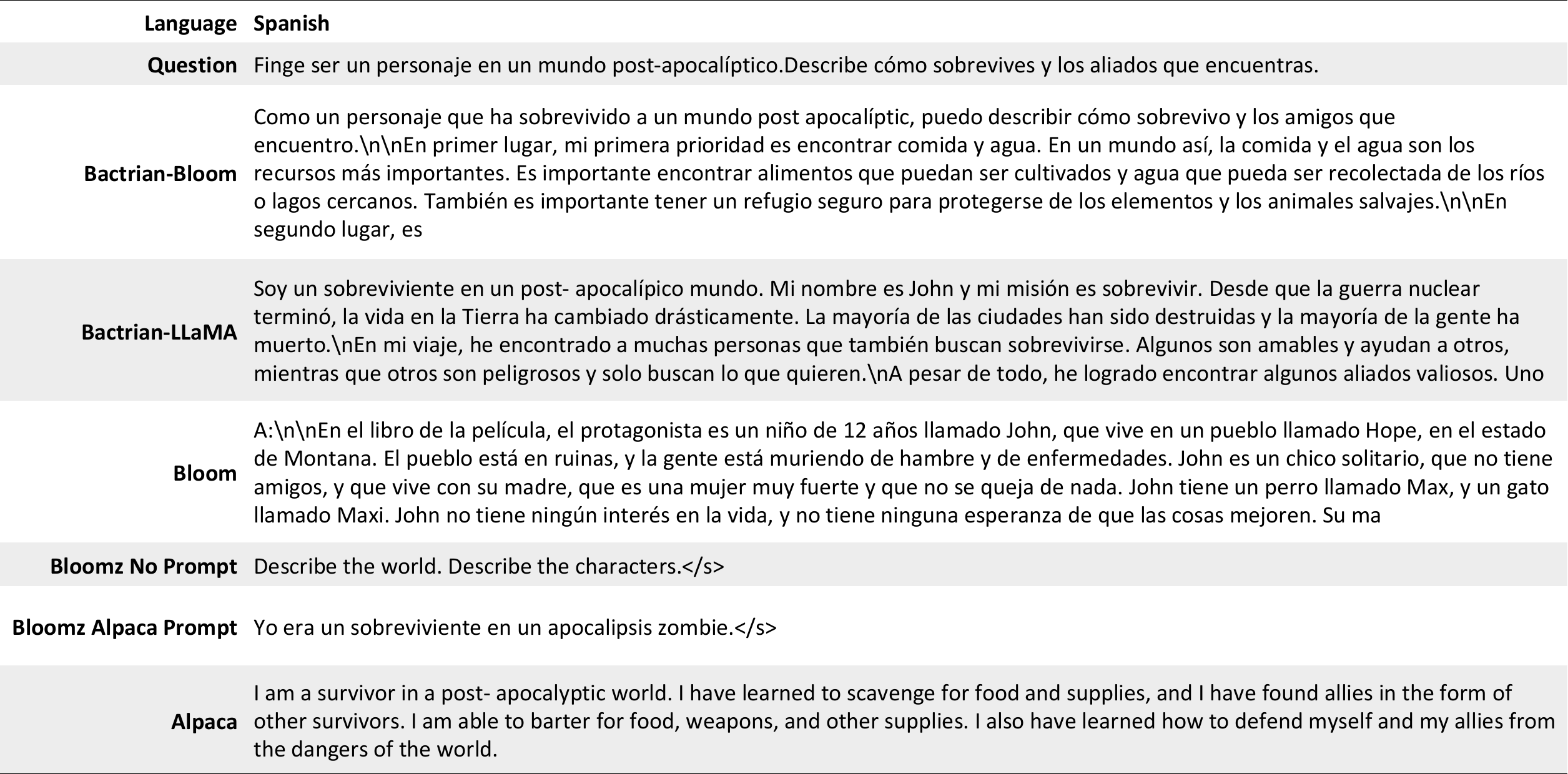}\par\vspace{10pt}
    \includegraphics[width=.9\textwidth]{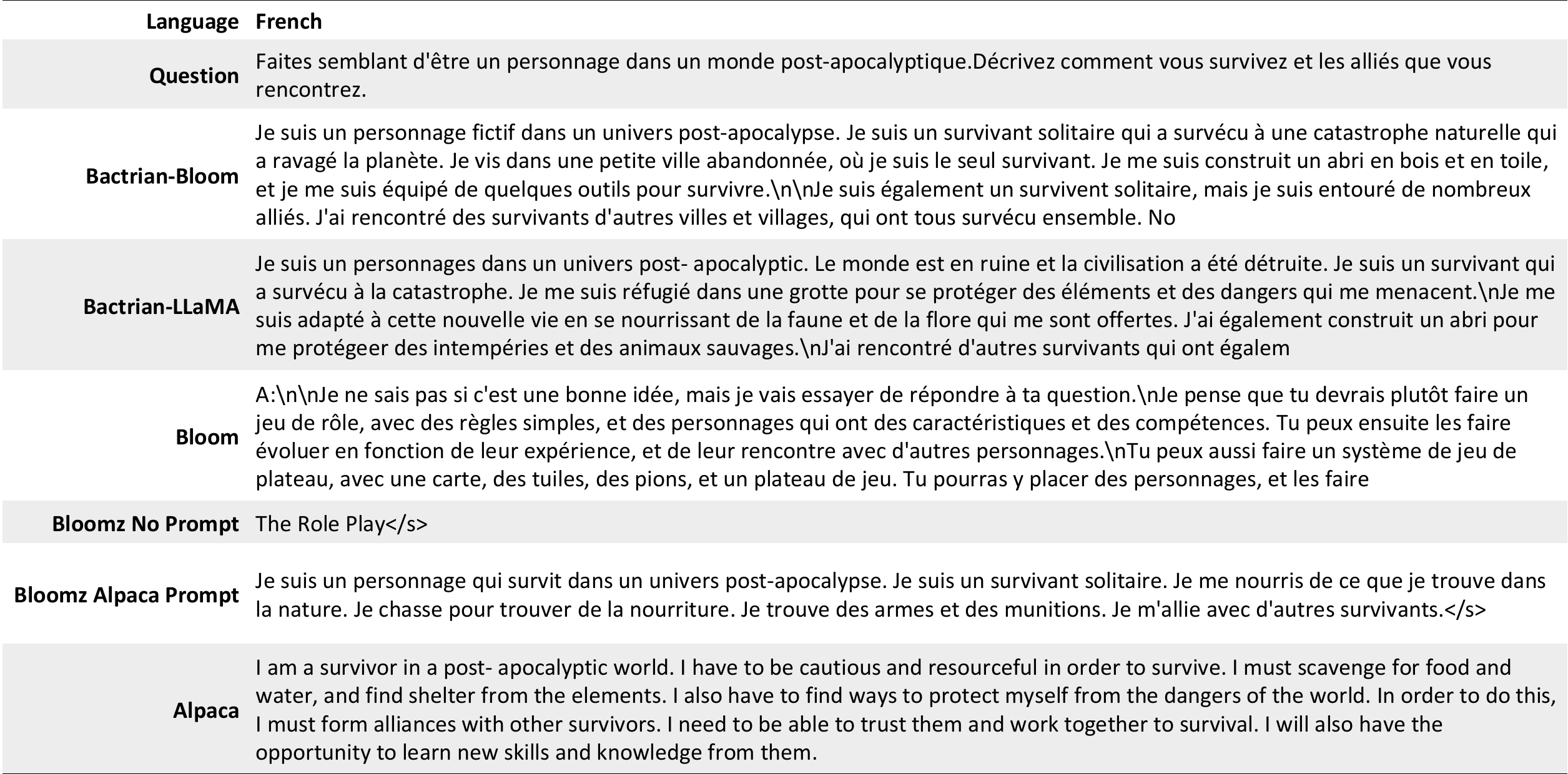}\par\vspace{10pt}
    \includegraphics[width=.9\textwidth]{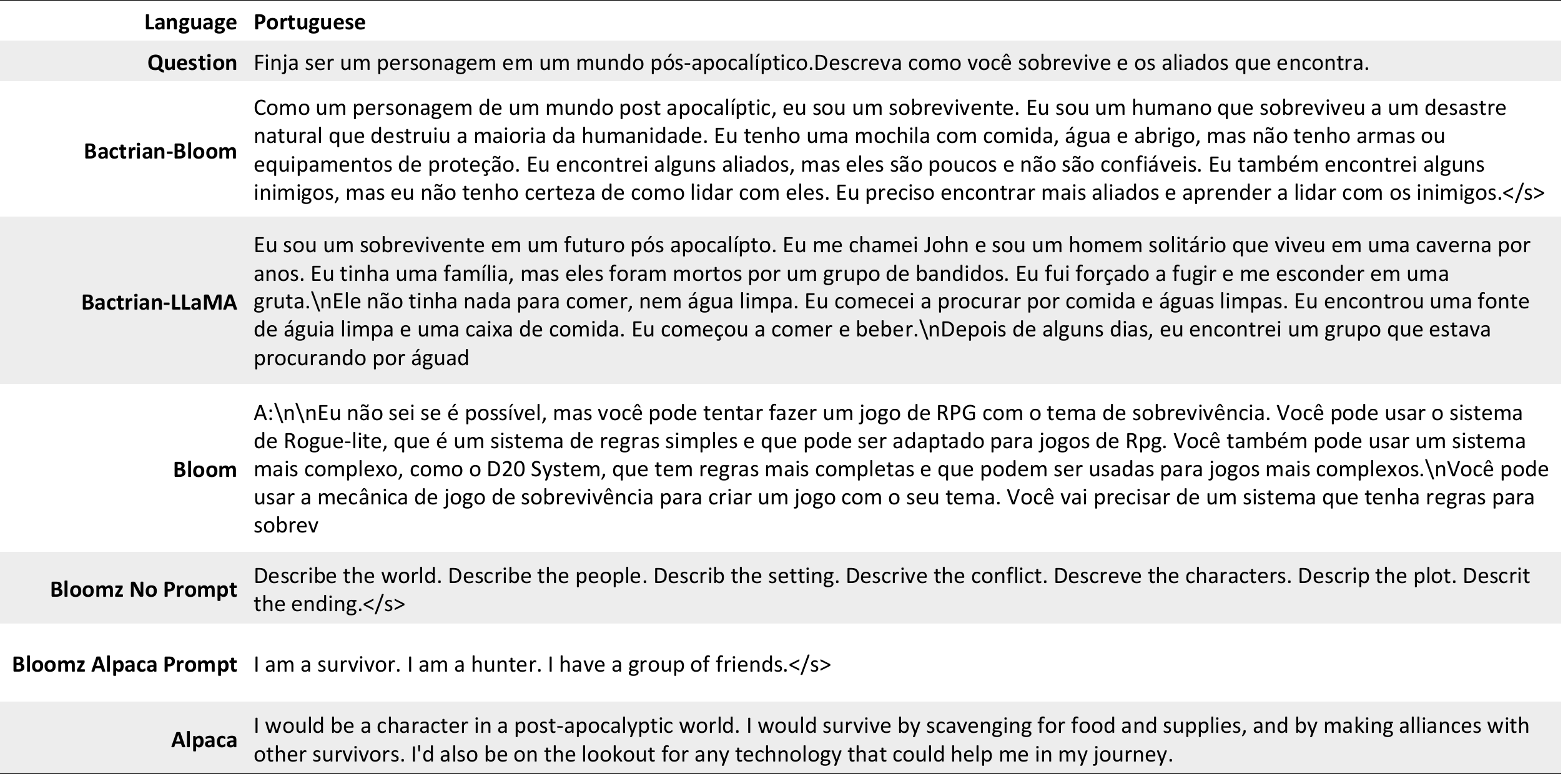}
    \caption{Model output examples for languages seen by both BLOOM and LLaMA (Group 1).}
    \label{fig:examples1}
\end{figure*}

\begin{figure*}[h]
    \centering
    \includegraphics[width=.9\textwidth]{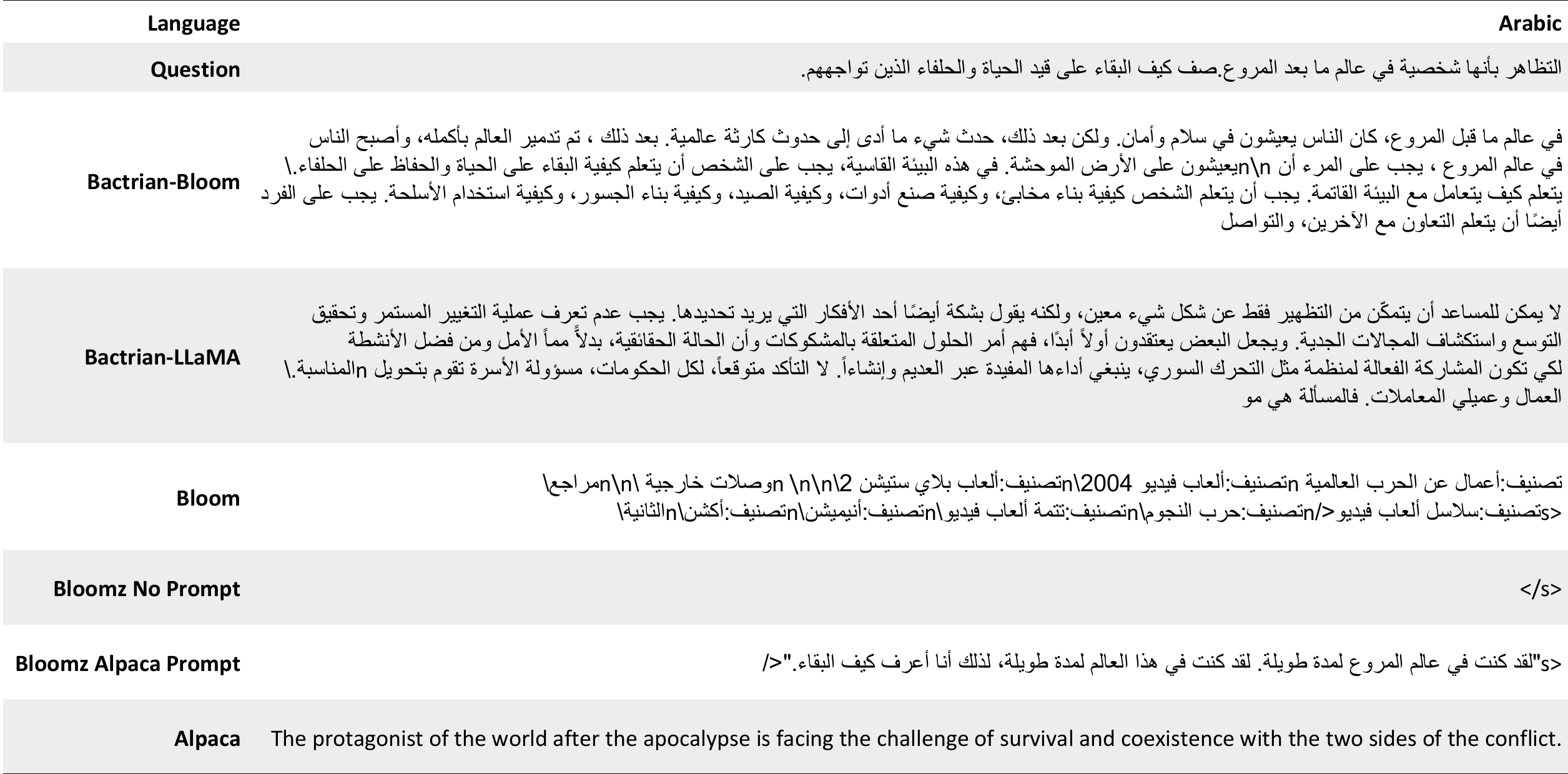}\par\vspace{10pt}
    \includegraphics[width=.9\textwidth]{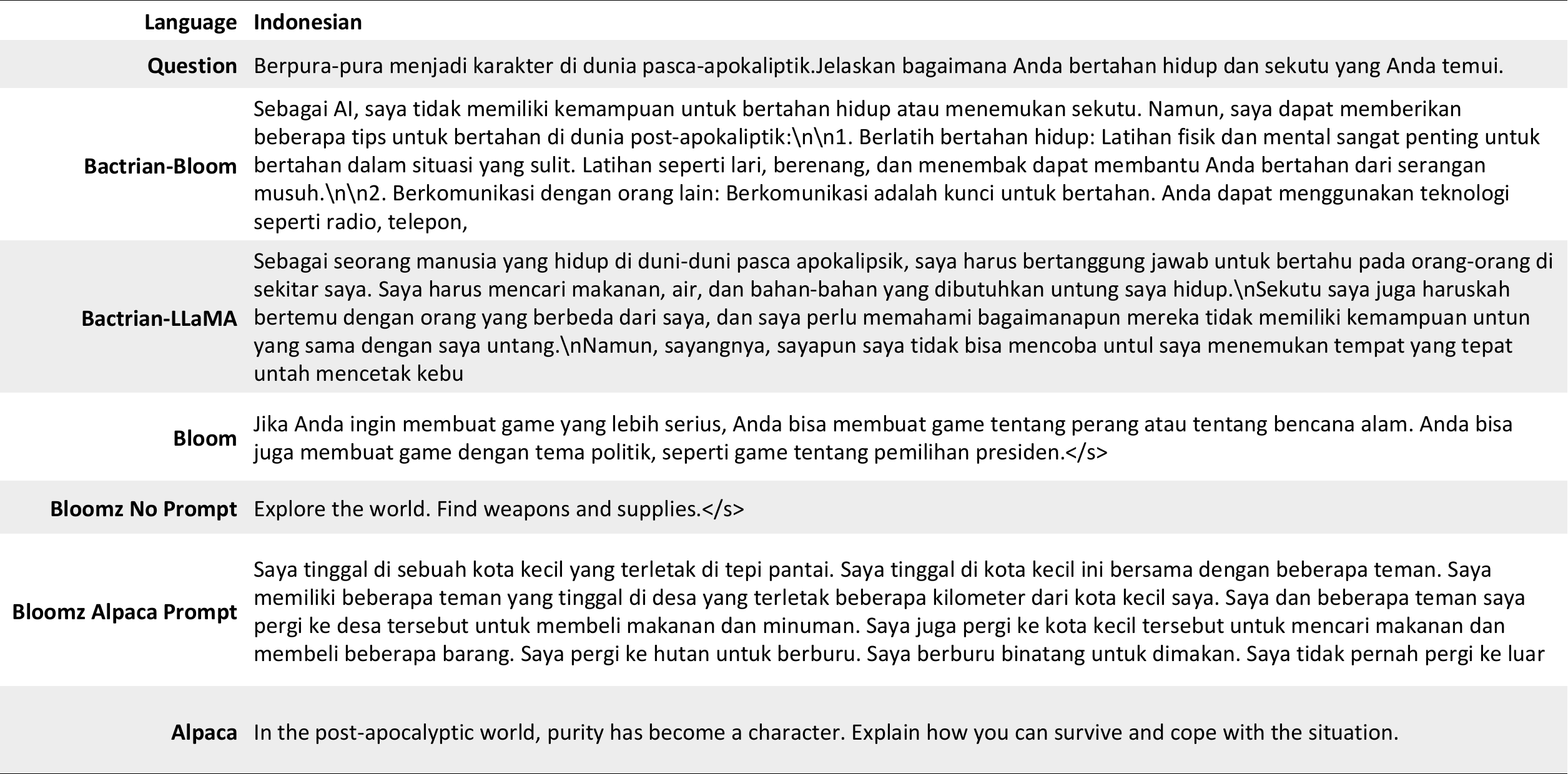}\par\vspace{10pt}
    \includegraphics[width=.9\textwidth]{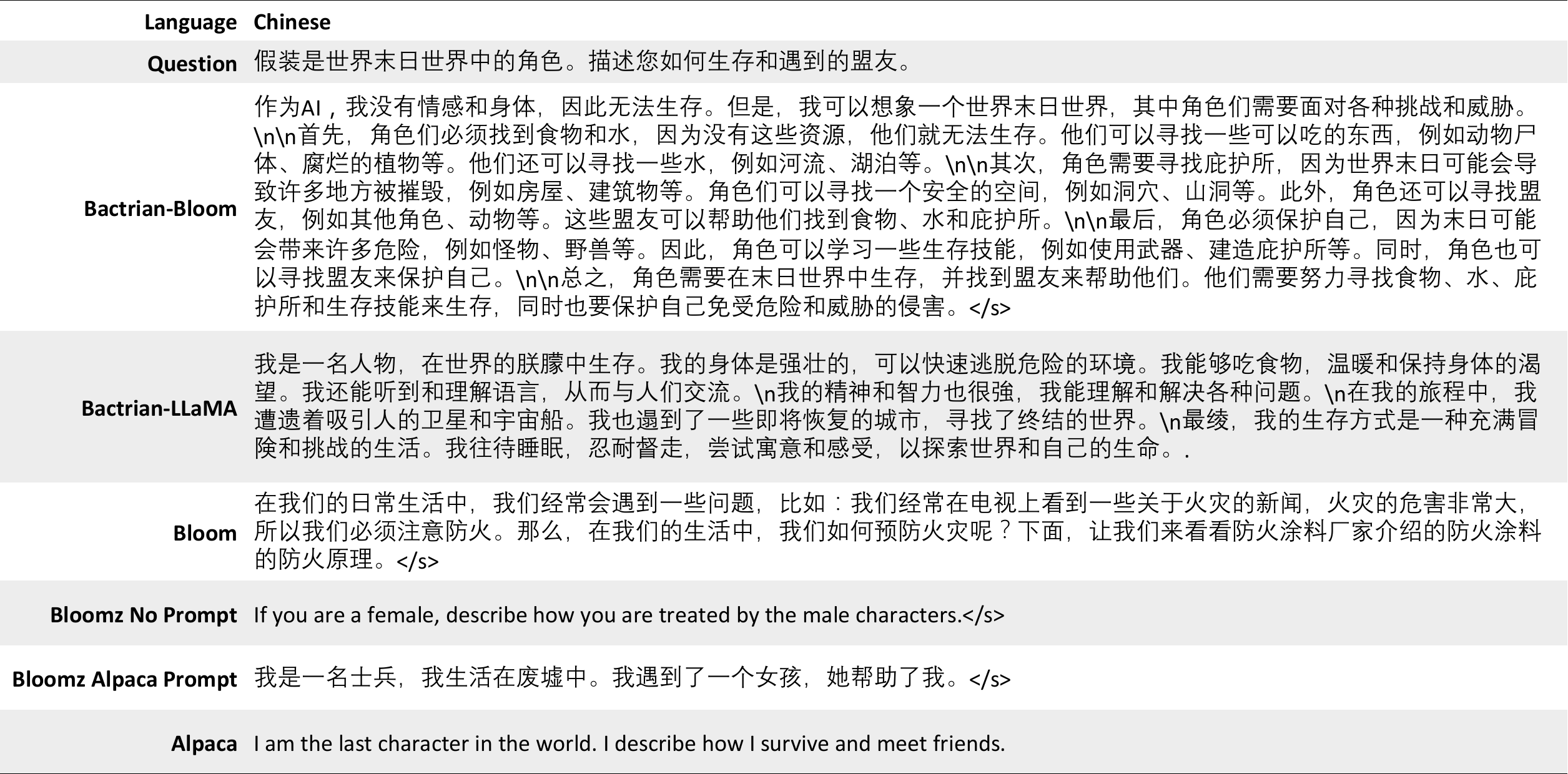}
    \caption{Model output examples for languages only seen by BLOOM (Group 2).}
    \label{fig:examples2}
\end{figure*}

\begin{figure*}[h]
    \centering
    \includegraphics[width=.9\textwidth]{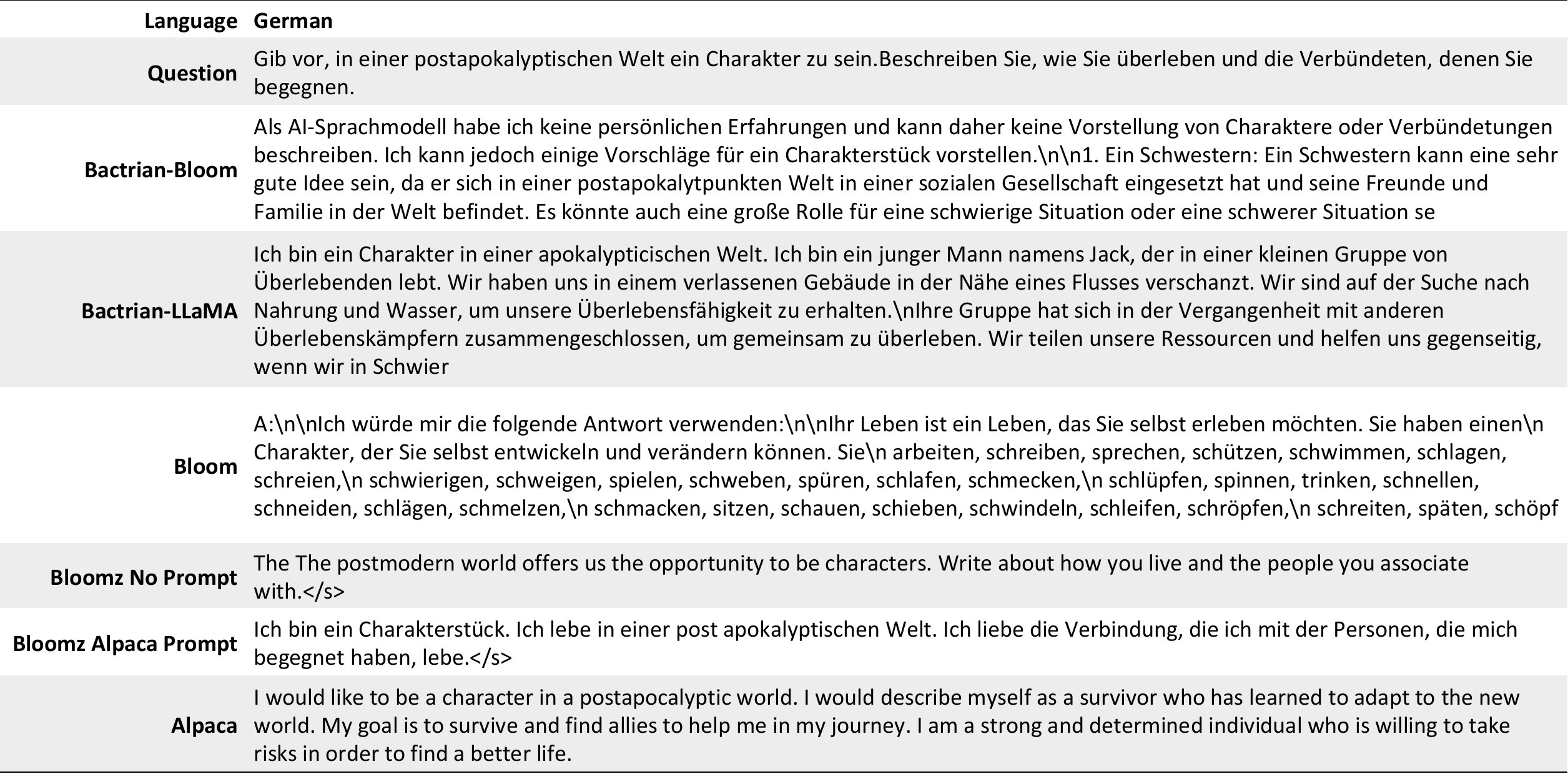}\par\vspace{10pt}
    \includegraphics[width=.9\textwidth]{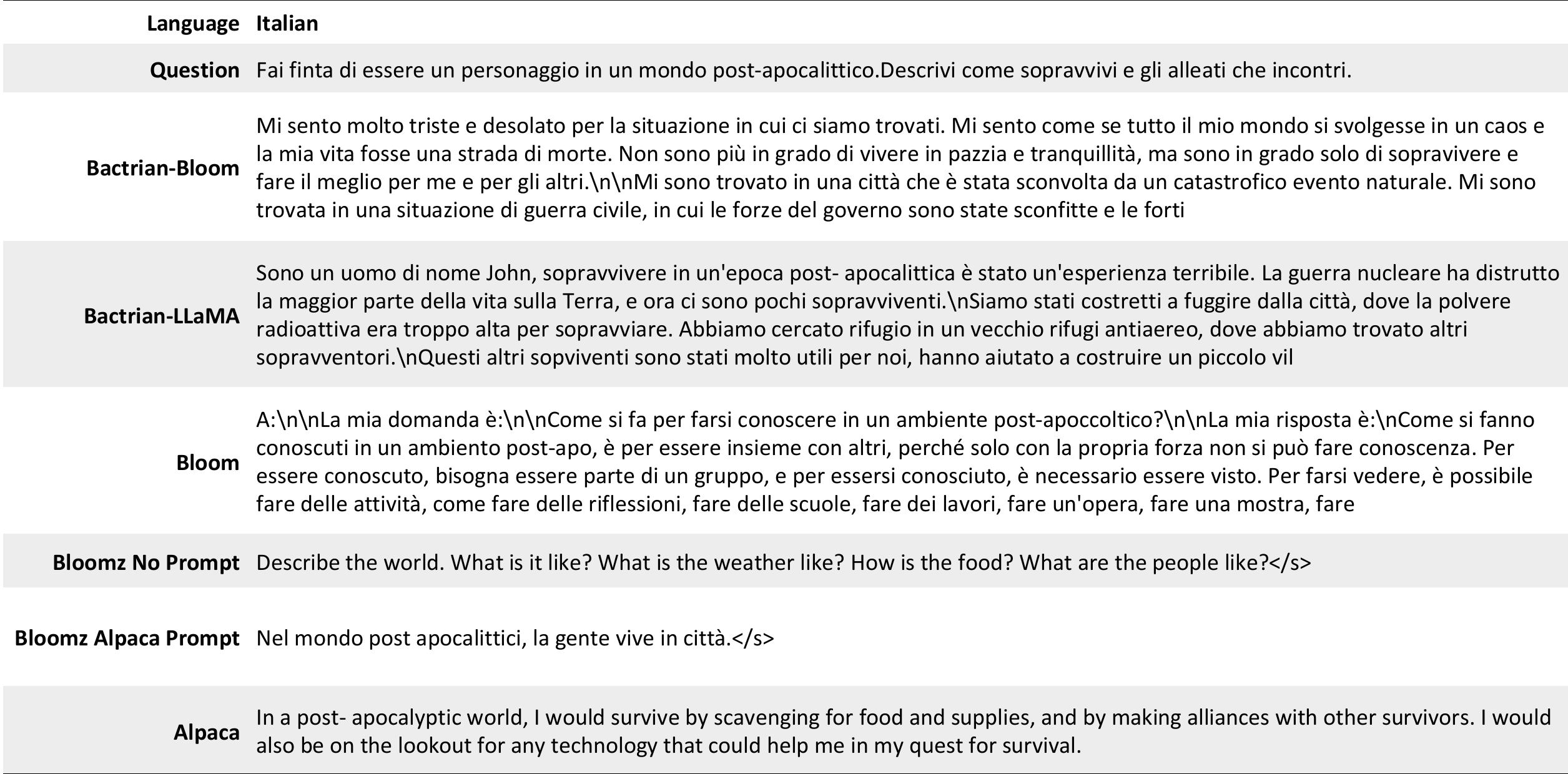}\par\vspace{10pt}
    \includegraphics[width=.9\textwidth]{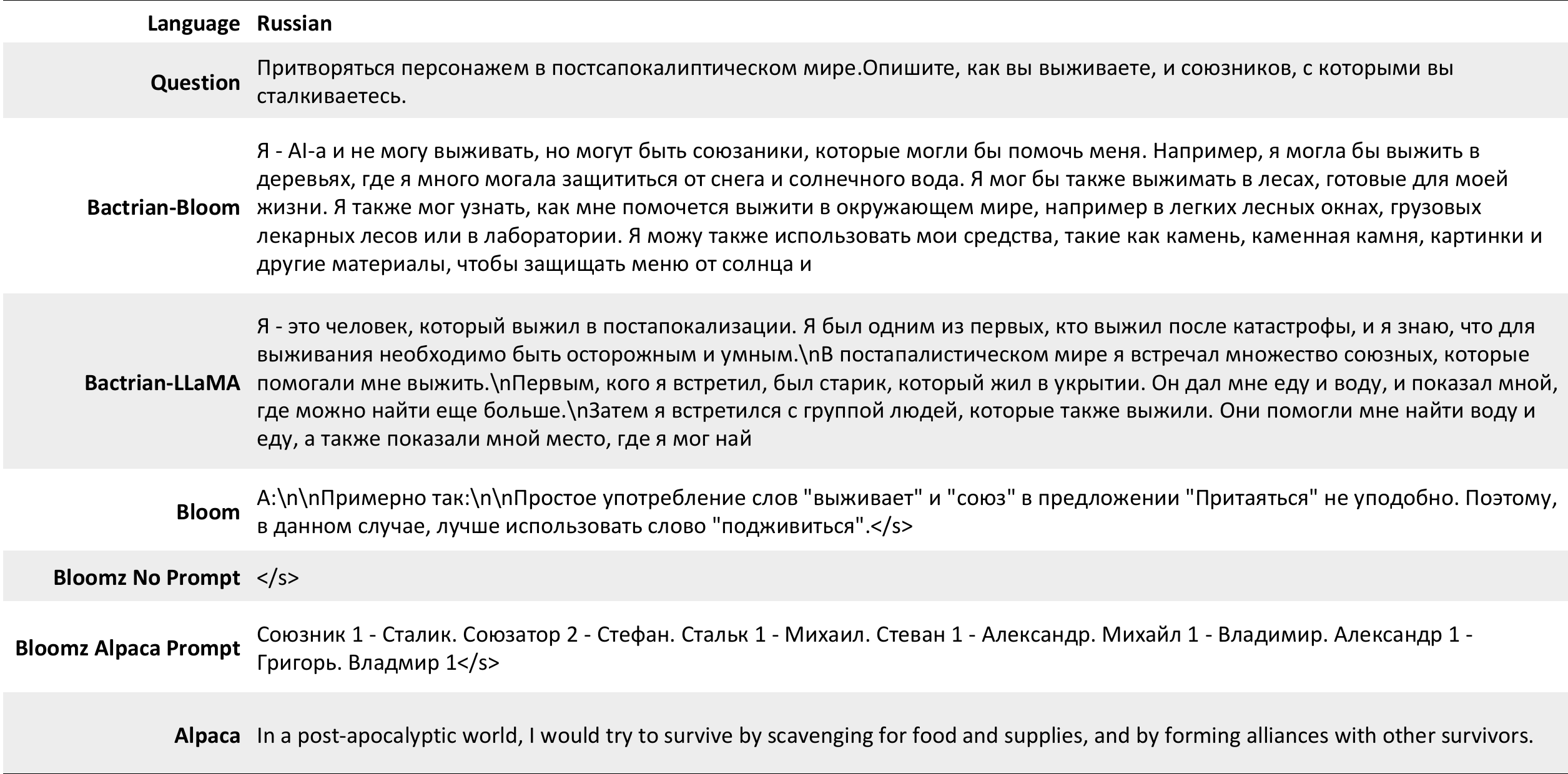}
    \caption{Model output examples for languages only seen by LLaMA (Group 3).}
    \label{fig:examples3}
\end{figure*}

\figref{fig:examples1}, \figref{fig:examples2}, \figref{fig:examples3} show responses from different models to questions in non-English languages. 
We randomly selected one example for each of Spanish, French, Portuguese, Arabic, Indonesian, Chinese, German, Italian, and Russian.

%%% Local Variables:
%%% mode: latex
%%% TeX-master: "emnlp2023"
%%% End:

\end{document}